\documentclass[letterpaper]{article} 
\usepackage{aaai24}  
\usepackage{times}  
\usepackage{helvet}  
\usepackage{courier}  
\usepackage[hyphens]{url}  
\usepackage{graphicx} 
\usepackage{natbib}  
\usepackage{caption} 
\usepackage{algorithm}
\usepackage{algorithmic}

\usepackage{newfloat}
\usepackage{listings}
\DeclareCaptionStyle{ruled}{labelfont=normalfont,labelsep=colon,strut=off} 
\floatstyle{ruled}
\newfloat{listing}{tb}{lst}{}
\floatname{listing}{Listing}
\nocopyright

\usepackage[utf8]{inputenc} 
\usepackage[T1]{fontenc}    

\PassOptionsToPackage{hyphens}{url}
\usepackage{booktabs}       
\usepackage{amsfonts}       
\usepackage{nicefrac}       
\usepackage{microtype}      
\usepackage[dvipsnames]{xcolor}

\usepackage{multirow}
\usepackage{graphicx}
\usepackage{amsmath}

\usepackage{amsthm}
\newtheorem{definition}{Definition}

\usepackage{xspace}
\newcommand{\protocol}{Rebalanced Deepfake Detection Protocol\xspace}
\newcommand{\protocolAbbrv}{RDDP\xspace}
\newcommand{\pcolconventional}{\textsc{conventional}\xspace}
\newcommand{\pcolwhite}{RDDP-\textsc{whitehat}\xspace}
\newcommand{\pcolsur}{RDDP-\textsc{surrogate}\xspace}
\newcommand{\model}{ID-Miner\xspace}

\newcommand{\ie}{i.e.}
\newcommand{\eg}{e.g.}

\title{In Anticipation of Perfect Deepfake: Identity-anchored Artifact-agnostic Detection under Rebalanced Deepfake Detection Protocol}
\author {
    Wei-Han Wang\textsuperscript{\rm 1},
    Chin-Yuan Yeh\textsuperscript{\rm 1,2},
    Hsi-Wen Chen\textsuperscript{\rm 1},
    De-Nian Yang\textsuperscript{\rm 2},
    Ming-Syan Chen\textsuperscript{\rm 1}
}
\affiliations {
    \textsuperscript{\rm 1}National Taiwan University
    \textsuperscript{\rm 2}Academia Sinica
    {\tt\small cyyeh@arbor.ee.ntu.edu.tw}
}

\begin{document}
\maketitle

\begin{abstract}
    As deep generative models advance, we anticipate deepfake videos achieving ``perfection''---exhibiting no discernible \emph{artifacts} or noise. However, current deepfake detectors, intentionally or inadvertently, rely on such artifacts for detection, as they are exclusive to deepfakes and absent in genuine examples. To bridge this gap, we introduce the \emph{\protocol (\protocolAbbrv)} to stress-test detectors under \emph{balanced} scenarios where genuine and forged examples bear \emph{similar} artifacts. We offer two \protocolAbbrv variants: \pcolwhite uses white-hat deepfake algorithms to create `self-deepfakes,' genuine portrait videos with the resemblance of the underlying identity, yet carry similar artifacts to deepfake videos; \pcolsur employs surrogate functions (\eg, Gaussian noise) to process both genuine and forged examples, introducing equivalent noise, thereby sidestepping the need of deepfake algorithms.

    Towards detecting perfect deepfake videos that aligns with genuine ones, we present \model, a detector that focus on extracting robust features anchored in the characteristic action sequences
    and disregards facile artifacts or appearances. Equipped with the \emph{artifact-agnostic loss} at frame-level and the \emph{identity-anchored loss} at video-level, \model effectively singles out identity signals amidst distracting variations. Extensive experiments comparing \model with $12$ baseline detectors under both conventional and \protocolAbbrv evaluations with two deepfake datasets, along with additional qualitative studies, affirm the superiority of our method and the necessity for detectors designed to counter perfect deepfakes.
\end{abstract}

\section{Introduction}

Deep generative models are capable of producing results nearly indistinguishable from real photos or videos~\cite{bond2021deep}. Unfortunately, highly realistic deepfake algorithms~\cite{siarohin2019first, doukas2021headgan, shu2022few, xu2022MobileFaceSwap} pose a severe threat of misinformation to the society through their fabrications~\cite{maddocks2020deepfake, westerlund2019emergence}. To counter the threat of deepfake, researchers have devoted significant efforts to propose various detection methods~\cite{rossler2019faceforensics++, chai2020makes, zhou2021joint}. However, to the best of our knowledge, \emph{all} deepfake detection methods rely on the distinct characteristics of deepfake videos, typically caused by the \emph{generative noise} or \emph{``artifacts''} created by the deepfake algorithms~\cite{wang2020cnn, zhou2021joint}. For instance, heuristic attributes found only in deepfake videos such as irregular eye blinking patterns~\cite{jung2020deepvision}, inconsistent head pose between inner and outer face regions~\cite{yang2019exposing}, or anomalies near the lips region~\cite{haliassos2021lips} have been leveraged, while end-to-end detections directly detect the distributional differences between deepfake and genuine videos~\cite{afchar2018mesonet, rossler2019faceforensics++}. While previous approaches typically perform well in detecting deepfake videos, overly depending on the anomalies in a deepfake video for detection is unreliable, since the continuous improvements in deep generative models may produce better algorithms that yield less irregularities.\footnote{For instance, preliminary study~\cite{corvi2023detection} shows that images created by diffusion models~\cite{dhariwal2021diffusion, ramesh2022hierarchical} have weaker artifacts that are more challenging to detect.} It is thus important to develop detection methods to reduce the dependence on identifying artifacts. Notably, recent research~\cite{agarwal2020detecting, idreveal2021cozzolino} has reframed deepfake detection as an \emph{identity-based detection problem}, reflecting concerns over artifact dependence. Under this approach, a genuine reference video of the individual portrayed in the analyzed video is provided during training and evaluation, enabling the detector to extract \emph{identity-based} features through pairwise comparisons. However, these works continue to operate under the conventional setting where an \emph{imbalance} of deepfake artifacts between forged and genuine examples allows their methods to rely on the such easily detected clues.

\begin{figure*}
    \centering
    \begin{minipage}[t]{0.325\textwidth}
         \centering
         \includegraphics[width=\textwidth]{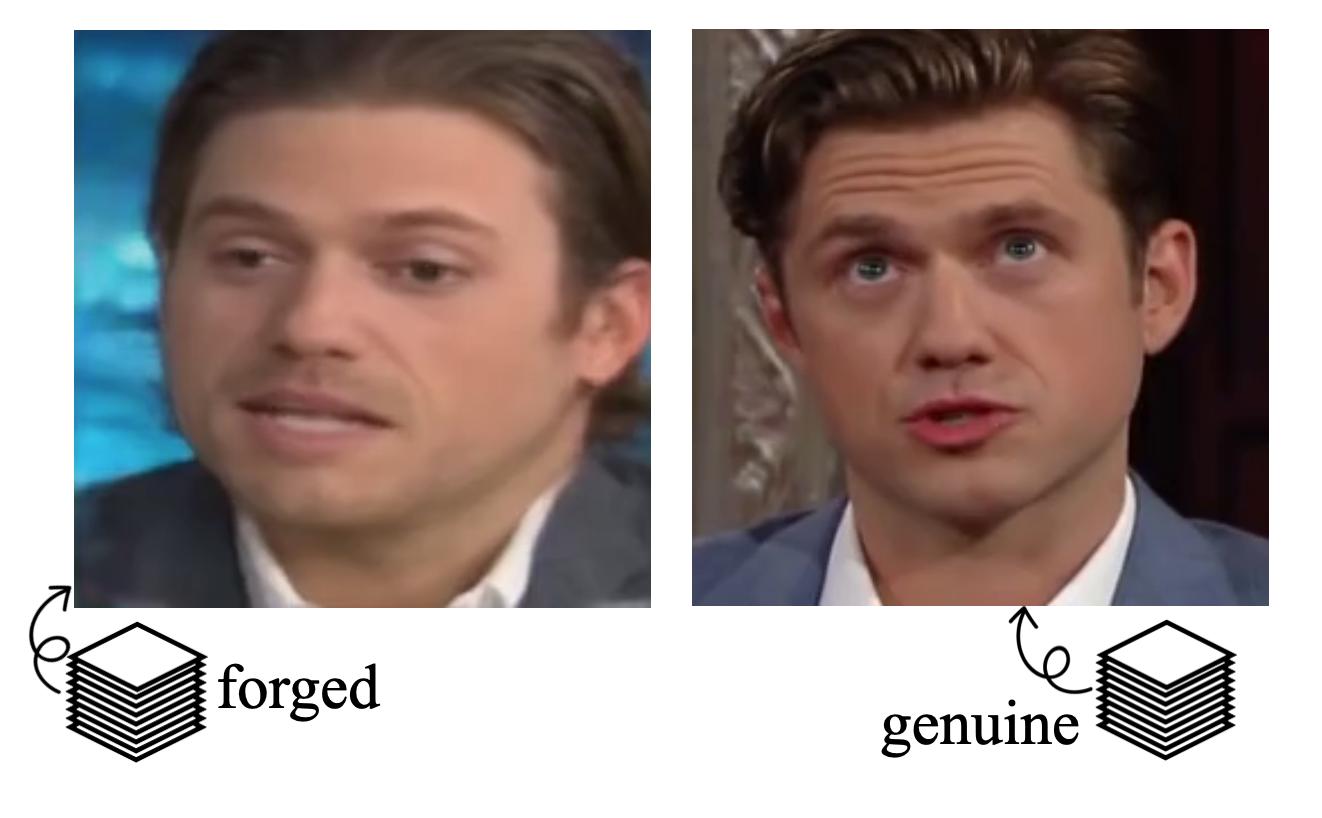}
         \pcolconventional
         \par
     \end{minipage}
     \hfill
     \begin{minipage}[t]{0.325\textwidth}
         \centering
         \includegraphics[width=\textwidth]{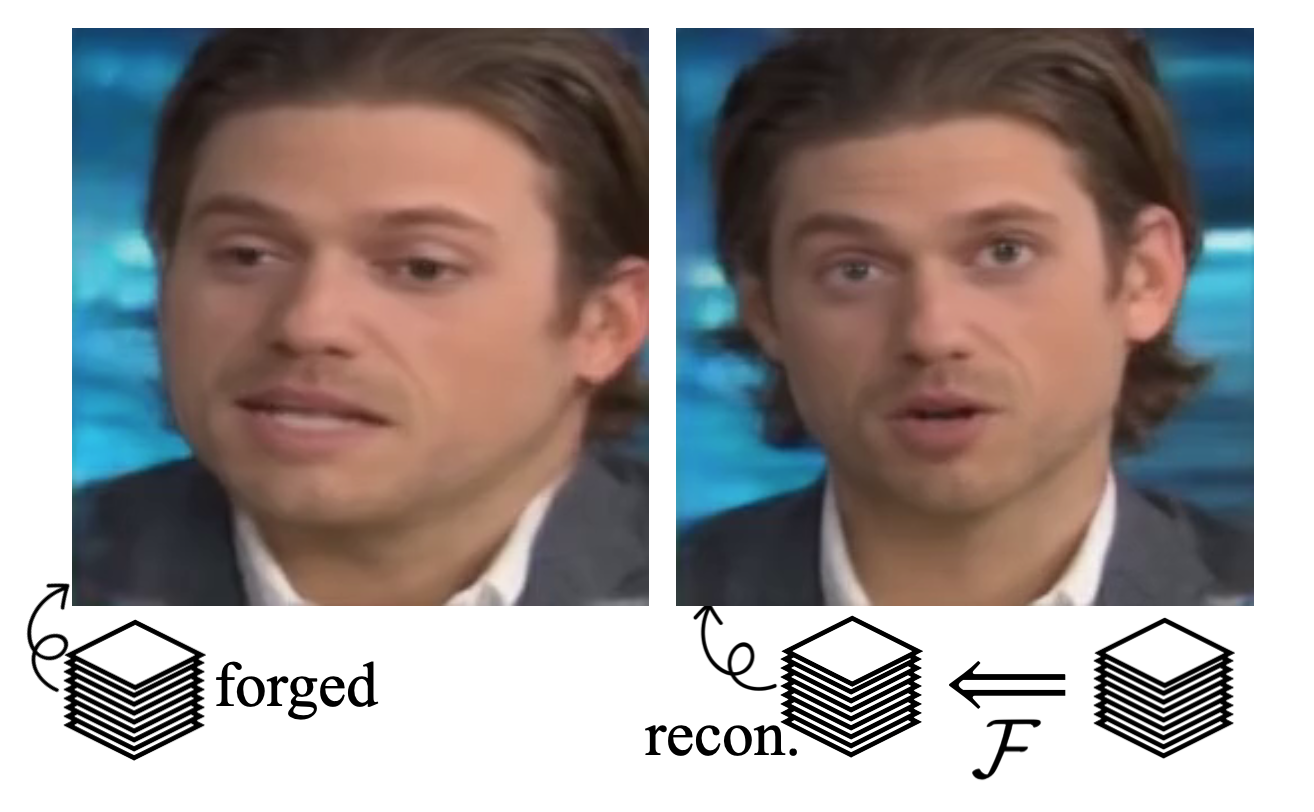}
         \pcolwhite
         \par
     \end{minipage}
     \hfill
     \begin{minipage}[t]{0.325\textwidth}
         \centering
         \includegraphics[width=\textwidth]{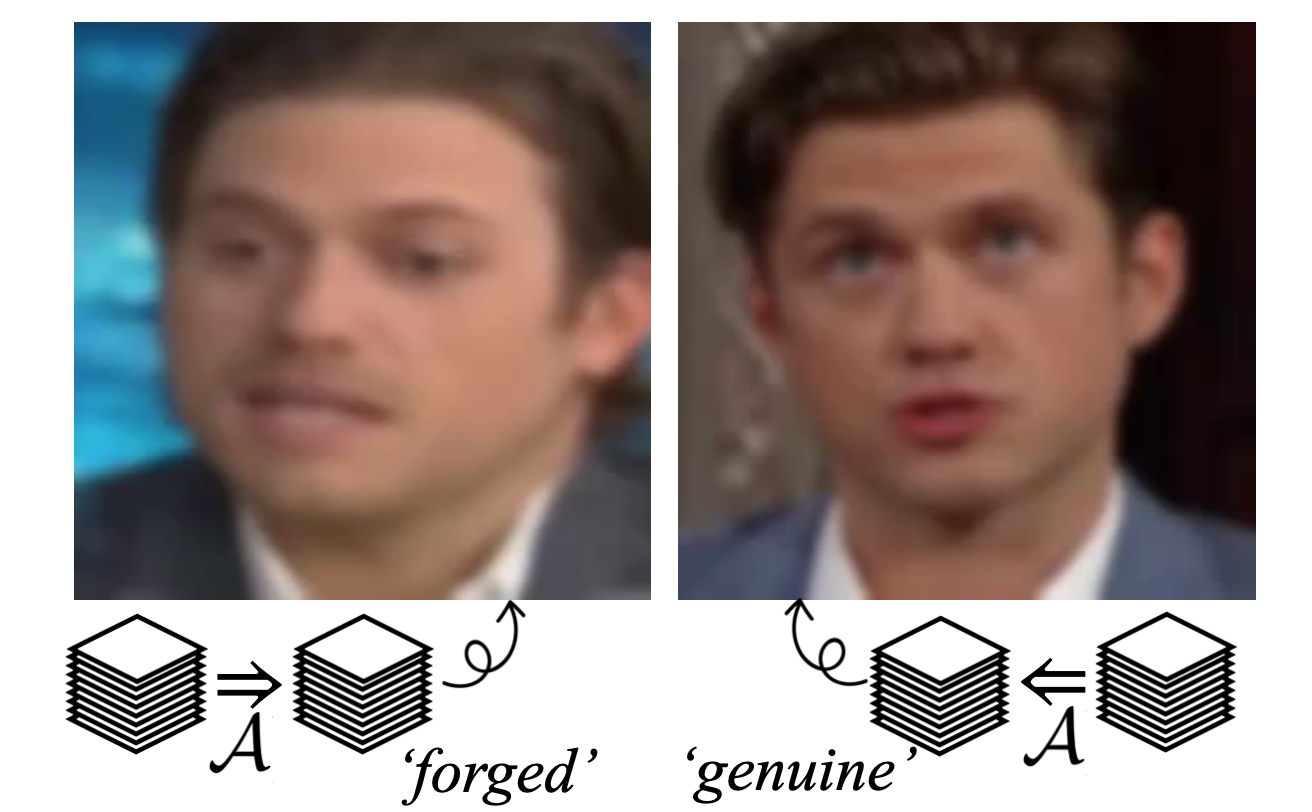}
         \pcolsur
         \par
     \end{minipage}
    \caption{\textbf{Comparison of the evaluation protocols.} The \pcolconventional protocol directly contrasts forged and genuine data, allowing deepfake detectors to exploit the distribution shift between the forged and the genuine examples. In contrast, \pcolwhite and \pcolsur reduce this bias by 1) processing genuine examples through a white-hat deepfake function ($\mathcal{F}$) to generate reconstructed examples (recon.), and 2) applying a surrogate function ($\mathcal{A}$) to both forged and genuine videos, respectively. By removing the obvious disparities between genuine and deepfake videos, \protocolAbbrv compels detectors to seek more robust detection cues, \eg, action sequences, as the above frame samples show that the differences in background and sharpness between forged and genuine in \pcolconventional is reduced in \pcolwhite by changing the genuine and in \pcolsur by transforming both sides uniformly.}
    \label{fig:compare-protocol}
\end{figure*}

In anticipation of future perfect deepfake, which may not contain artifacts and become almost perfectly aligned with the distribution of genuine videos, it becomes increasingly important to design an evaluation framework that could test deepfake detectors under a setting where deepfake and genuine videos are indifferentiable from mere appearances. Inspired by adversarial training~\cite{madry2017towards} and image augmentation techniques~\cite{perez2017effectiveness}, we first observe that it is possible to add the same type of deepfake artifacts onto the genuine videos, or to add a different type of noise onto both data, (\ie, Gaussian noise), for reducing the disparity between `true' and `fake' examples. It enables us to compel the detection methods to identify more robust and critical attributes, thereby providing a more stringent test. 

Furthermore, we discern that current deepfake algorithms do not exhibits the ability to generate novel facial actions, but must source the action from a given input video~\cite{siarohin2019first}. Therefore, our idea is to let the detection model concentrate less on the appearance and artifacts but focus more on the patterns rooted in the action sequence of each person because past research such as \emph{gait detection}~\cite{pappas2001reliable} has demonstrated that human walking behaviour contains identifiable characteristics due to habits or biological differences. While such idea have not been applied to deepfake detection, we argue that a person's identity can also be determined for a given portrait video based on the sequence of facial actions.

In this work, we propose the \emph{\protocol (\protocolAbbrv)}, an evaluation framework for artifact-independent deepfake detection. We design two \protocolAbbrv variations. \pcolwhite leverages a white-hat deepfake algorithm to \emph{reconstruct} examples, in order to directly imbue deepfake-specific artifacts into the genuine videos. By contrast, \pcolsur applies surrogate functions such as resize, JPEG compression, video compression, and Gaussian blur to induce a consistent noise in both genuine and forged examples, to overshadow existing disparities between the two. In particular, we reconstruct genuine videos into \emph{self-deepfakes} in \pcolwhite by using them as the driving video which provides the action sequences, to manipulate the same person's face as the appearance. As a result, these \emph{self-deepfakes} has the same facial action sequences \emph{and} the same appearance of the original genuine video, yet also contains deepfake artifacts. 

In addition, to achieve a true identity-based detection method that could still function under more difficult \protocolAbbrv evaluations, we propose \emph{Identity-anchored Artifact-agnostic Deepfake Detection (\model)}. \model subscribes to the above principles underlying \protocolAbbrv and learns to identify the puppeteer behind the appearance of a deepfake forgery by ignoring the artifacts and concentrating on the \emph{action sequences}. In particular, \model comprises 1) a pre-trained deep learning-based Facial Action Unit (FAU) extractor~\cite{baltrusaitis2018openface}, followed by 2) an \emph{artifact-agnostic encoder} at the frame level, and 3) an \emph{identity-anchored aggregator}
to process the frame-level embeddings at the video level. We design and employ contrastive losses~\cite{chen2020simple} at both levels to ensure an \emph{identity-anchored, artifact-agnostic} detection. The \emph{artifact-agnostic loss} at the frame level guides the encoder to derive consistent embeddings from image frames, regardless of the presence of artifacts. At the video level, we introduce the \emph{identity-anchored loss}, which emphasizes the subject's action sequences over their appearances. In particular, different forgeries with identical action sequences are considered as similar examples, while those with differing actions are treated as dissimilar, even if they share the same face. Therefore, \model learns to discover identifiable characteristics based on the action sequences, rather than appearances or deepfake-induced artifacts. As a result, \model learns to find consistent identity features that persists over deepfake algorithm modifications. In contrast, prior works focus on locating deepfake-specific features, and would be fooled when such features are removed from future more advanced deepfakes.


Our contributions are summarized as follows: \textbf{1)} We introduce \emph{\protocol (\protocolAbbrv)}, which quantitatively demonstrates the significant performance degradation in baseline detectors due to over-reliance on the distinctive imperfections contained solely in deepfake videos. \textbf{2)} We present two \protocolAbbrv variants including \pcolwhite and \pcolsur, to create a more stringent test with or without a ``white-hat'' deepfake algorithm. \textbf{3)} \model, equipped with the frame-level \emph{artifact-agnostic loss} and the video-level \emph{identity-anchored loss}, outperforms $12$ baseline detectors under \protocolAbbrv and maintains substantial effectiveness in the conventional setting. \textbf{4)} We further demonstrate the robustness and generalizability of \model through quantitative experiments across three evaluation protocols, two datasets, and against $12$ baselines, as well as qualitative analyses that further support the claimed functionality of both \protocolAbbrv and \model.

\section{Related Work}

Deepfake techniques~\cite{westerlund2019emergence}, starting with faceswap (FS)~\cite{deepfake2017faceswap} and advancing to Face Reenactment (FR)~\cite{siarohin2019first}, enables anyone to manipulate other person's appearances due to the widespread accessibility of open-sourced projects~\cite{xu2022MobileFaceSwap, iperov2018deepfacelab, siarohin2019first}, posing significant societal risks. To counter this, researchers actively develop detection strategies~\cite{zi2020wilddeepfake}, yet they all exploit the consistent deficiencies in current deepfake outputs~\cite{wang2020cnn, zhou2021joint}. Initial approaches used binary classification for end-to-end training~\cite{chollet2017xception, afchar2018mesonet, nguyen2019capsule, rossler2019faceforensics++}. Subsequent efforts targeted features such as eye blinking~\cite{li2018ictu, jung2020deepvision}, face boundary imperfections~\cite{li2018exposing, li2020face, zhao2021learning, shiohara2022detecting} or inconsistencies between inner and outer face regions~\cite{agarwal2020detecting, dong2022protecting}. Others address textural artifacts~\cite{zhao2021multi}, frequency domain patterns~\cite{li2021frequency, qian2020thinking}, temporal inconsistencies~\cite{zheng2021exploring, liu2023ti2net}, or noise from up-sampling~\cite{wang2020cnn, durall2020watch, liu2021spatial}. However, as revealed by our \protocolAbbrv evaluations, these methods' reliance on artifacts created by current deepfake algorithms significantly limit their effectiveness against better deepfakes in the future.

Recently, identity-based detection pivot towards identifying features associated with individual identities. For instance, \citet{agarwal2019protecting} craft separate models specialized for each target, while later works compare video embeddings' similarity to reference videos of the target identity~\cite{agarwal2020detecting, idreveal2021cozzolino}. Despite these advances, their effectiveness diminishes in \protocolAbbrv because they still operate under unbalanced settings with marked differences between forged and genuine examples. This insight informs the design of \model, which effectively extract representations based on action sequences instead of appearances or artifacts that could be influenced by deepfake.

Another line of studies also considers generalization~\cite{guan2022delving, chen2022ost} and audio features incorporation~\cite{ji2021audio, agarwal2023audio}, though it remains in the conventional, unbalanced settings. Similarly, some research claims robust deepfake detection. FrePGAN~\cite{jeong2022frepgan} improves the deepfake detector's robustness by adding perturbations onto deepfake videos. However, while they mitigate the problem of overfitting on specific types of artifacts, they did not address the risk of \emph{relying} on artifacts. In contrast, \protocolAbbrv directly reduces the distributional difference for a more stringent evaluation, while \model excels under \protocolAbbrv by adopting an identity-anchored approach and focusing on the robust action-sequences-based features. Besides, counteractive measures such as adversarial noise~\cite{yeh2020disrupting, yeh2021attack} or hidden watermarks~\cite{asnani2022proactive, zhao2023proactive} have been proposed, yet require preemptive actions against deepfake forgeries and could not protect the victims once deepfake is created and spread. In contrast, our focus is on stress-testing and advancing deepfake detection, essential in counteracting the spread of deepfake. 

\begin{figure*}
    \centering
    \includegraphics[width=.36\linewidth]{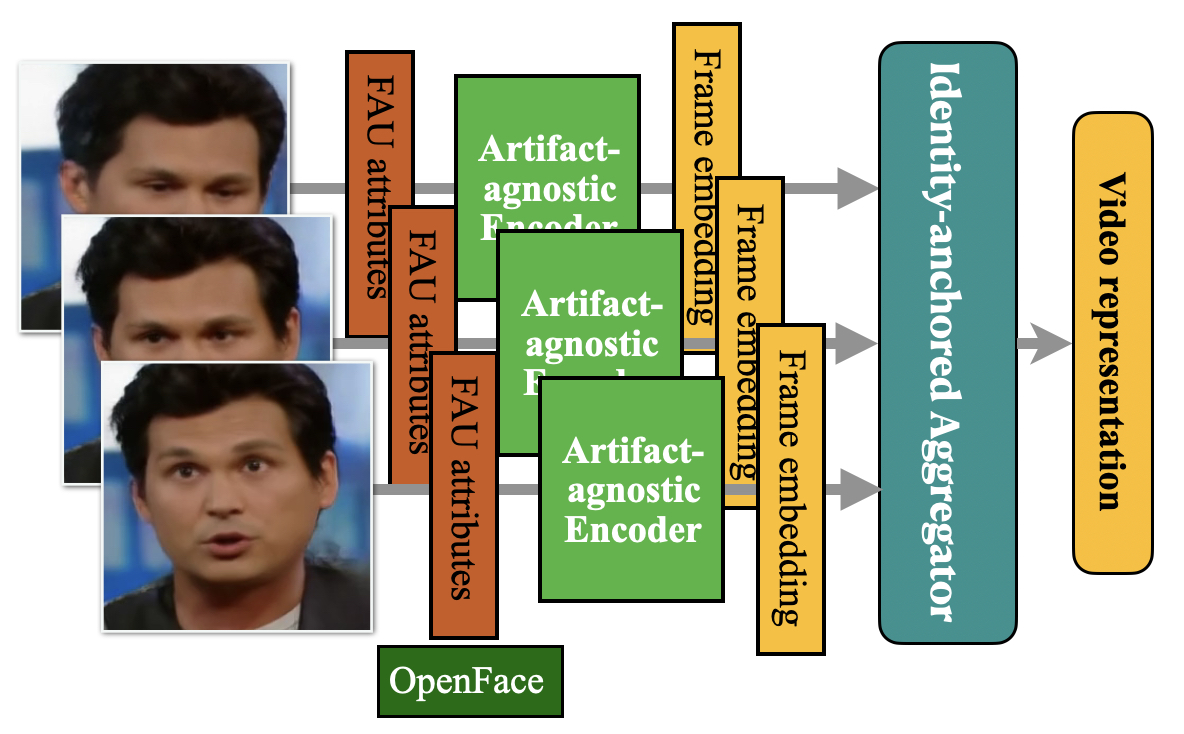}
    \includegraphics[width=.63\linewidth]{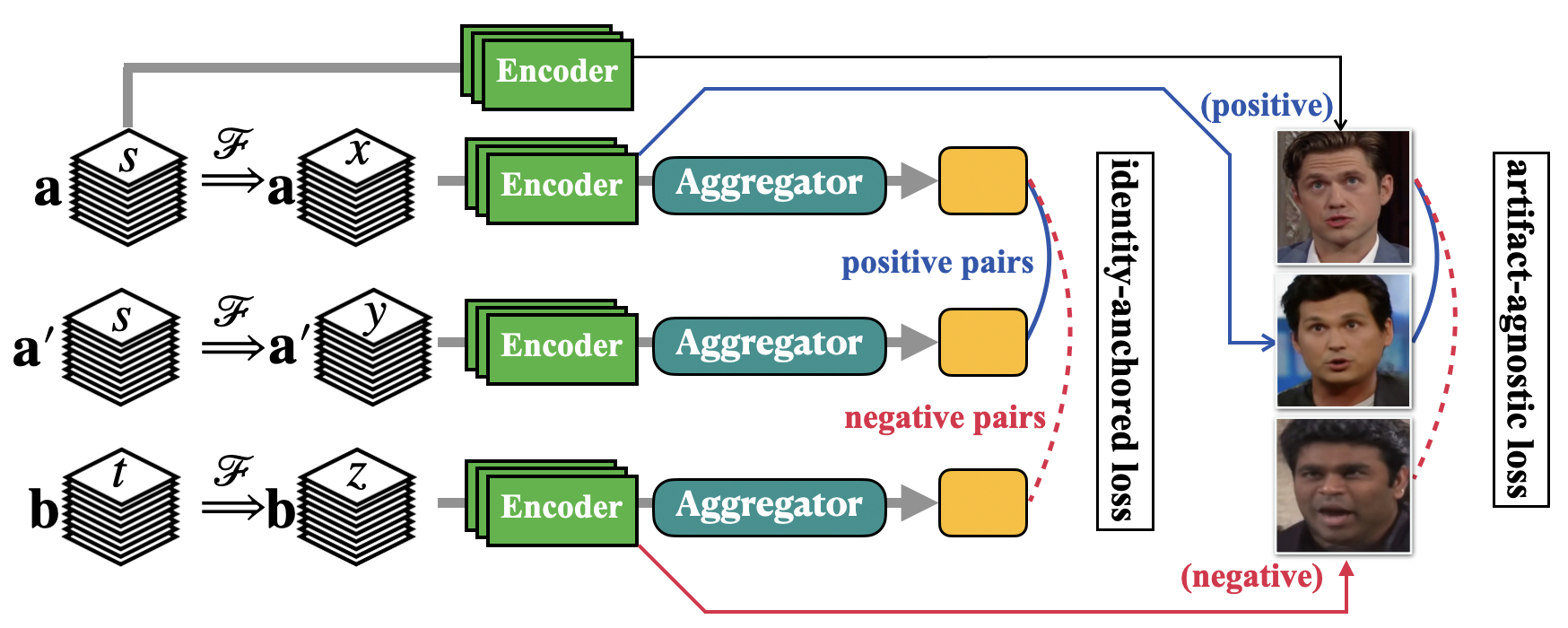}
    \caption{\textbf{The architecture and training of \model.} \model employs a hierarchical structure to extract embeddings with frame-level encoders from Facial Action Unit (FAU) attributes, then process the embeddings into a video-level representation. During training, the \emph{identity-anchored loss} discriminates between video representations of the same individual's action sequences ($\mathbf{a}$ and $\mathbf{a'}$) and those of a different individual ($\mathbf{b}$), irrespective of the facial appearances ($x$, $y$, or $z$). Concurrently, the frame-level encoder is trained by the \emph{artifact-agnostic loss} to sample frames from videos pre- and post-deepfake transformation to prioritize encoding facial expressions and actions over artifacts.}
    \label{fig:idminer}
\end{figure*}

\section{Approach}
\subsection{Problem formulation}
\label{subsec:problem}

We focus our attention to deepfakes revolving personal portrait videos. In particular, a video of subject $s$ performing facial action sequence $\mathbf{a}$ is denoted as $V(s, \mathbf{a})$, with either component omitted when context allows. The $i^{th}$ frame $V_i$ displays an image of subject $s$ engaged in facial action $\mathbf{a}_i$. Given a target subject $x$ and a driving video $V(s, \mathbf{a})$, deepfake algorithms may be expressed as a mapping $\mathcal{F}_x\circ V(s, \mathbf{a})\mapsto U(x, \mathbf{a})$, where $\circ$ and subscript $x$ denote the input of video $V$ and a portrait image of $x$ to the deepfake algorithm $\mathcal{F}$, respectively. Various methods have been devised to detect deepfakes, yet they all operate under a conventional framework that directly contrasts deepfake and genuine videos and their effectiveness against \emph{perfect} deepfakes could not be inferred. Thus, we introduce \protocolAbbrv along with its two variants. In the following, we define the different approaches for the deepfake detection problem.

\begin{definition}[\pcolconventional protocol]
    We denote a set of genuine videos as $D_{gen} = \{V\}$ and a set of forged videos as $D_{forg} = \{U\mid U= \mathcal{F}_x\circ V(s), V(s)\in D_{gen}, x\neq s\}$, where $x$ represents a different identity to $s$. The conventional protocol differentiate $D_{gen}$ and $D_{forg}$\emph{~\cite{afchar2018mesonet, idreveal2021cozzolino, liu2023ti2net}}.
\end{definition}

As current deepfake algorithms are imperfect, a significant distribution shift appears between $D_{forg}$ and $D_{gen}$ and is thereby easily detected. In anticipation of future deepfake algorithms which would not have such imperfections, we introduce the \protocolAbbrv to reduce the distribution shift. 

\begin{definition}[\pcolwhite]
    Given $D_{gen}$ and $D_{forg}$, \pcolwhite uses a white-hat deepfake $\mathcal{F}'$ to instill deepfake artifacts into $D_{gen}$ and creates \emph{reconstructed examples} $D_{recon} =
    \{V'\mid V'=\mathcal{F}'_s\circ V(s),\,\,\forall V\in D_{gen}\}$, where each video $V'$ is reconstructed by $\mathcal{F}'$ using the portrait image of the same person ($s$). Under \pcolwhite, the goal is to differentiate between $D_{recon}$ and $D_{forg}$.
\end{definition}

While \pcolwhite effectively equalizes the testing environment, its use of white-hat deepfake may not only raise ethical concerns but also pose challenges in terms of practical implementation. Therefore, we present another idea to exploit surrogate functions which introduce noise into the processed videos, as an alternative. Following prior work~\cite{jiang2020deeperforensics, zheng2021exploring, dong2022protecting}, we select four challenging real-world perturbations for detection methods as our surrogate functions $\mathcal{A}$, including resize (\ie, a consecutive down- and up-sampling), JPEG compression, video compression, and adding Gaussian blur.

\begin{definition}[\pcolsur]
    Given a noise-inducing surrogate function $\mathcal{A}$, 
    we apply $\mathcal{A}$ to both the forged set $D_{forg}$ and the genuine set $D_{gen}$ to introduce identical noise into both sets. The goal is to differentiate the noise-added sets $\tilde{D}_{gen} = \mathcal{A}\circ D_{gen}$ and $\tilde{D}_{forg} = \mathcal{A}\circ D_{forg}$.
\end{definition}

Note that for identity-based detector evaluations, genuine videos $\hat{V}(s)$, reconstructed videos $\mathcal{F}'_s\circ \hat{V}(s)$, and surrogate-processed videos $\mathcal{A}\circ \hat{V}(s)$ are provided as reference under \pcolconventional, \pcolwhite, and \pcolsur, respectively.

\subsection{\model}
\label{subsec:model}

Fig.~\ref{fig:idminer} (left) displays the architecture of \model, utilizing a hierarchical process to derive representation vectors from videos. The model focuses on ``mining'' identifiable information from a subject's action sequence while effectively ignoring artifacts and appearances. Consistent with \pcolwhite, white-hat deepfake algorithm are used to recreate the genuine videos, aligning the distribution between genuine and deepfake instances by introducing similar artifacts. Specifically, action sequences are harvested from genuine videos and replicated using an image of the same subject, ensuring a consistent action and appearance with the original video. By employing these recreated samples, we formulate the artifact-agnostic loss at the frame level and the identity-anchored loss at the video level (see Fig.~\ref{fig:idminer} (right)). In contrast to prior works' attention to deepfake imperfections, these loss functions intentionally restricts \model from leveraging the facile visual artifacts and redirects its focus towards directly mining the action-based features consistent across real and fake examples, bolstering \model's ability to handle the more difficult \protocolAbbrv evaluations.



\paragraph{Frame-level embedding process.} To extract action-sequence based features, we first employ OpenFace~\cite{baltrusaitis2018openface}, a facial behavior analysis toolkit, to obtain attribute vectors of Facial Action Units (FAU)~\cite{ekman1978facial} for each frame. We then process these vectors with an \emph{artifact-agnostic encoder} to generate frame-level embeddings. We devise the \emph{artifact-agnostic loss} $\mathcal{L}_{artifact}$, a contrastive loss designed to ensure consistent embeddings for image frames with and without artifacts. With the above frame-level embedding process denoted as $\mathcal{E}$,
\begin{equation}\small
\label{eqn:artifact-agnostic-loss}
    \mathcal{L}_{artifact} = - \log\frac{e^{\mathbf{q}_i \cdot \mathbf{k}^+_i/\tau}}{e^{\mathbf{q}_i \cdot \mathbf{k}_i^+/\tau} + \sum e^{\mathbf{q}_i\cdot\mathbf{k}_j^-/\tau}},
\end{equation}
\begin{equation}\small
    \text{with} \left[\begin{array}{ll}
    \mathbf{q}_i \in \mathcal{E}\circ V(s,\mathbf{a})_i\\
    \mathbf{k}^+_i \in \mathcal{E}\circ \mathcal{F}_x\circ V(s,\mathbf{a})_i\\
    \mathbf{k}^-_j \in \mathcal{E}\circ\mathcal{F}_z\circ V(t,\mathbf{b})_j
    \end{array}\right],
\end{equation}
where subscripts $i$ and $j$ indicate different frames, $\mathcal{F}$ is a white-hat deepfake algorithm, the dot notation ($\cdot$) signifies cosine similarity, $\tau$ is the temperature parameter, and $\sum$ denotes sampling over negative examples $\mathbf{k}^-_j$ within the batch. Intuitively, $\mathcal{L}_{artifact}$ aims to ensure that frame embeddings remain consistent for the same images pre- and post-deepfake processing, while retaining essential features that indicate distinct facial actions, thereby causing \model to be agnostic towards artifacts at the frame-level.

\paragraph{Video-level representation aggregation.} To aggregate the frame-level embeddings of a video, we employ Gated Recurrent Units (GRU)~\cite{cho2014learning} for an efficient design in our \emph{identity-anchored aggregator}. Specifically, the aggregator processes each frame embedding sequentially to generate a full-video representation. We introduce the \emph{identity-anchored loss} $\mathcal{L}_{identity}$ to converge the representations of videos with action sequences originating from the same person and to separate those from different identities. Denoting the complete \model model as a function $\mathcal{M}$, 
\begin{equation}
\label{eqn:identity-anchored-loss}
    \mathcal{L}_{identity} = - \log\frac{e^{\mathbf{q}\cdot \mathbf{k}^+/\tau}}{e^{\mathbf{q}\cdot \mathbf{k}^+/\tau} + \sum e^{\mathbf{q}\cdot\mathbf{k}^-/\tau}},
\end{equation}
\begin{equation}\small
    \text{with} \left[\begin{array}{ll}
    \mathbf{q} \in \mathcal{M}\circ\mathcal{F}_x\circ V(s, \mathbf{a})\\
    \mathbf{k}^+ \in \mathcal{M}\circ\mathcal{F}_y\circ V(s, \mathbf{a}')\\
    \mathbf{k}^- \in \mathcal{M}\circ\mathcal{F}_z\circ V(t, \mathbf{b})
    \end{array}\right],
\end{equation}
    where $x$ and $y$ indicate different individuals and $z$ being an arbitrary identity, potentially equating to $x$. $\mathbf{a}$ and $\mathbf{a}'$ represent action sequences pertaining to the same person ($s$), while $\mathbf{b}$ denotes that of a different individual ($t$). $\mathcal{L}_{identity}$ anchors the full video representation to each identity, ensuring that the video representations of \model are consistent across diverse action sequences from the same individual, regardless of the appearance. Given that the frame-level embeddings are artifact-agnostic, this advantage extends to the video-level representation as well.

\paragraph{Training procedure.} We jointly train both losses by combining them into the total loss as
\begin{equation}\small
    \label{eqn:total-loss}
    \mathcal{L}_{total} = \mathcal{L}_{identity} + \lambda\mathcal{L}_{artifact},
\end{equation}
where $\lambda$ balances the identity-anchored loss and the artifact-agnostic loss during the training phase. In practice, we prepare the primary training data batches based on the positive and negative video pairs required by the \emph{identity-anchored} loss, then randomly retrieve frames from selected genuine and deepfake-forged videos in the training batches to derive the \emph{artifact-agnostic} loss (see Fig.~\ref{fig:idminer}).

\paragraph{Identification procedure.} Following~\cite{idreveal2021cozzolino}, we provide a reference video for each test video under examination, according to the video subject's \emph{appearance} (face). With \model, we extract the video representation vectors for both videos and employ the cosine similarity to assess the consistency between the action sequence of the video and the identity of the depicted face.

\section{Experiment}

\subsection{Experiment Setup}
\subsubsection{Datasets.} Our experiments are based on large-scale public portrait video datasets, including VoxCeleb~\cite{nagrani2017voxceleb} with over $20$k videos across $1251$ subjects, and Celeb-DF~\cite{li2020celeb} consisting of $590$ genuine videos, $5639$ deepfake videos over $59$ subjects. We establish two divisions of detection evaluations corresponding to face reenactment (FR) and faceswap (FS). For FR, we select genuine examples from VoxCeleb, then utilize the First Order Motion Model~\cite{siarohin2019first} to generate the forged and the reconstructed examples. For FS, we use genuine and forged examples from Celeb-DF while generating reconstructed samples from the genuine examples with MobileFaceSwap~\cite{xu2022MobileFaceSwap}.

\subsubsection{Implementation.} 
All video examples are portrait aligned, cropped, and resized to $256\times 256$. We train \model with the Adam optimizer~\cite{kingma2015adam} ($\beta_1 = 0.9$ and $\beta_2 = 0.999$) for 150 epochs with $\lambda = 0.1$ and $\tau = 0.07$. Leveraging its exceptional generalizability, (see Table~\ref{tab:generalizability}) we train \model solely in FR to test in both FR and FS. Following~\cite{guan2022delving, chen2022ost, idreveal2021cozzolino}, we use the Area Under the Receiver Operating Characteristic Curve (AUC) as evaluation metrics. A pair of NVIDIA RTX 3080 and 3090 GPUs are used in our experiments.

\subsection{Conventional and \protocolAbbrv evaluations}
\begin{table*}[t]\small
    \caption{\textbf{Detection evaluation for face reenactment (FR).} \emph{avg. drop} shows the average performance reduction in \protocolAbbrv relative to \pcolconventional. \model delivers exceptional results under all \protocolAbbrv evaluations while maintaining competitive performance under the conventional setting. \model (no FLE) represents the ablation of \model with no frame-level encoder.}
    \label{tab:full-compare-FR}
    \centering
    \resizebox{0.95\textwidth}{!}{
    \begin{tabular}{l c c cccc c}
        \toprule
        & \multirow{3}{*}{\shortstack[l]{\textsc{conven}\\ -\textsc{tional}}} 
        & \multirow{3}{*}{\shortstack[l]{RDDP-\\\textsc
        {whitehat}}} 
        & \multicolumn{4}{c}{\pcolsur} 
        & \multirow{2}{*}{\shortstack[l]{\\ \textit{avg. drop}}} \\
        \cmidrule(lr){4-7} 
        &                   & & Resize & JPEG & Video Compression & Gaussian Blur \\
        
        \midrule
        Xception~\cite{chollet2017xception} & $0.916$ & $0.605$ & $0.681$ & $0.584$ & $0.621$ & $0.561$ & $-33.4\%$ \\
        MesoNet~\cite{afchar2018mesonet} & $0.922$ & $0.626$ & $0.670$ & $0.568$ & $0.573$ & $0.546$ & $-35.3\%$ \\
        EfficientNet~\cite{tan2019efficientnet} & $\mathbf{0.948}$ & $0.760$ & $0.758$ & $0.788$ & $0.781$ & $0.724$ & $-19.6\%$  \\
        FTCN~\cite{zheng2021exploring} & $\underline{0.925}$ & $0.679$ & $0.795$ & $0.790$ & $0.722$ & $0.734$ & $-19.6\%$ \\
        TI2Net~\cite{liu2023ti2net} & $0.905$ & $0.662$ & $\underline{0.808}$ & $0.703$ & $0.680$ & $0.667$ & $-22.2\%$ \\
        \midrule
        PWL~\cite{agarwal2019protecting} & $0.893$ & $0.679$ & $0.689$ & $0.678$ & $0.699$ & $0.690$ & $-23.1\%$ \\
        A\&B~\cite{agarwal2020detecting} & $0.624$ & $0.577$ & $0.565$ & $0.566$ & $0.600$ & $0.595$ & $\underline{-7.0\%}$ \\
        ID-Reveal~\cite{idreveal2021cozzolino} & $0.743$ & $0.566$ & $0.670$ & $0.599$ & $0.577$ & $0.531$ & $-20.8\%$ \\
        \midrule
        \model & $0.876$ & $\mathbf{0.837}$ & $\mathbf{0.833}$ & $\mathbf{0.847}$ & $\mathbf{0.849}$ & $\mathbf{0.749}$ & $\mathbf{-6.1\%}$  \\
        \model (no FLE) & $0.898$ & $\underline{0.795}$ & $0.800$ & $\underline{0.840}$ & $\underline{0.813}$ & $\underline{0.738}$ & $-11.2\%$ \\
        \bottomrule
    \end{tabular}
    }
\end{table*}

\begin{table*}[t]\small
    \caption{\textbf{Detection evaluation for faceswap (FS).} 
    Note that \model is trained under FR without further finetuning, yet still yields consistently good performances under FS.
    }
    \label{tab:full-compare-FS}
    \centering
    \resizebox{0.95\textwidth}{!}{
    \begin{tabular}{l c c cccc c}
        \toprule
        & \multirow{3}{*}{\shortstack[l]{\textsc{conven}\\ -\textsc{tional}}} 
        & \multirow{3}{*}{\shortstack[l]{RDDP-\\\textsc
        {whitehat}}} 
        & \multicolumn{4}{c}{\pcolsur} 
        & \multirow{2}{*}{\shortstack[l]{\\ \textit{avg. drop}}} \\
        \cmidrule(lr){4-7} 
        &                   & & Resize & JPEG & Video Compression & Gaussian Blur \\

        \midrule
    Xception~\cite{chollet2017xception} & $\mathbf{0.912}$ & $0.654$ & $0.693$ & $0.606$ & $0.640$ & $0.587$ & $-30.3\%$ \\
    MesoNet~\cite{afchar2018mesonet} & $0.797$ & $0.544$ & $0.523$ & $0.486$ & $0.491$ & $0.508$ & $-36.0\%$ \\
    FWA~\cite{li2018exposing} & $0.640$ & $0.547$ & $0.582$ & $0.559$ & $0.561$ & $0.520$ & $-13.5\%$ \\
    EfficientNet~\cite{tan2019efficientnet} & $0.901$ & $0.698$ & $0.727$ & $0.686$ & $0.667$ & $0.655$ & $-23.8\%$ \\
    Face X-ray~\cite{li2020face} & $0.877$ & $0.654$ & $0.685$ & $0.621$ & $0.591$ & $0.572$ & $-28.8\%$ \\
    FTCN~\cite{zheng2021exploring} & $0.882$ & $0.628$ & $\underline{0.799}$ & $0.706$ & $0.729$ & $\mathbf{0.776}$ & $-17.5\%$ \\
    EFNB4+SBIs~\cite{shiohara2022detecting} & $\underline{0.911}$ & $0.685$ & $0.743$ & $0.709$ & $0.679$ & $0.595$ & $-25.1\%$ \\
    ICT\cite{dong2022protecting} & $0.847$ & $0.611$ & $0.793$ & $0.784$ & $0.790$ & $0.627$ & $-14.9\%$  \\
    TI2Net~\cite{liu2023ti2net} & $0.877$ & $0.618$ & $0.765$ & $0.656$ & $0.635$ & $0.623$ & $-24.8\%$ \\
    \midrule
    PWL~\cite{agarwal2019protecting} & $0.878$ & $0.668$ & $0.653$ & $0.655$ & $0.688$ & $0.684$ & $-23.7\%$ \\
    A\&B~\cite{agarwal2020detecting} & $0.569$ & $0.517$ & $0.504$ & $0.506$ & $0.539$ & $0.534$ & $\underline{-8.6\%}$ \\
    ID-Reveal~\cite{idreveal2021cozzolino} & $0.811$ & $0.573$ & $0.712$ & $0.618$ & $0.588$ & $0.596$ & $-23.9\%$ \\
    \midrule
    \model & $0.859$ & $\mathbf{0.823}$ & $\mathbf{0.820}$ & $\mathbf{0.832}$ & $\mathbf{0.834}$ & $\underline{0.743}$ & $\mathbf{-5.7\%}$  \\
    \model (no FLE) & $0.896$ & $\underline{0.770}$ & $0.781$ & $\underline{0.826}$ & $\underline{0.795}$ & $0.708$ & $-13.4\%$ \\
        \bottomrule
    \end{tabular}
    }
\end{table*}

To anticipate perfect deepfakes which emit no distinct generative noise, deepfake detection methods must not rely on such artificial features. Nevertheless, we observe a significant performance drop across all baselines when confronted with the ``rebalanced'' \protocolAbbrv evaluations where both forged and genuine class samples contain similar artifacts, indicating a dependency on the original distribution difference. Table~\ref{tab:full-compare-FR} and Table~\ref{tab:full-compare-FS} present the detection performances under all protocols for FR and FS, respectively. As shown in both tables, while most baseline methods display commendable performance in the \pcolconventional setting, they substantially decline under \protocolAbbrv evaluations. For instance, in Table~\ref{tab:full-compare-FR}, all non-identity-based detection methods (top $5$ rows) achieve AUC scores exceeding $0.9$ in the \textsc{conventional} setting for FR. However, their performances constantly face a drop of $18\%$ to $35\%$ when subjected to \protocolAbbrv evaluations. Identity-based detections (middle $3$ rows) also face significant performance decline under \protocolAbbrv. Results in Table~\ref{tab:full-compare-FS} show a more dramatic decline for the baseline methods. Several AUC scores dropped to close to $0.5$ in the \protocolAbbrv evaluations, which is barely better than random chance. In contrast, our proposed \model suffers a slighter decrease and consistently outperforms the baseline in both \pcolwhite and \pcolsur evaluations, confirming its robustness against these challenging conditions. Our \model learns to differentiate and identify portrait videos based on robust action sequences under the identity-anchored, artifact-agnostic training. Thus, in the \protocolAbbrv evaluations where the visual artifacts are less useful, our method surpasses the baselines by a significant margin. 


\paragraph{Ablation study.} We compare the performance of ID-Miner with its ablated version, \model (no FLE), which removes the frame-level artifact-agnostic encoder and directly aggregates the FAU attribute vectors with the identity-anchored aggregator. The results, presented in the final two rows of Table~\ref{tab:full-compare-FR} and Table~\ref{tab:full-compare-FS}, indicate that excluding the \emph{artifact-agnostic encoder} bolsters the performances in \pcolconventional yet leads to a more substantial degradation in RDDP evaluations. This comparison highlights the important role of frame-level \emph{artifact-agnostic loss} in reducing artifact dependency. Nevertheless, the ablated version still outperforms all baseline methods in $7$ of the $10$ \protocolAbbrv evaluations between Table~\ref{tab:full-compare-FR} and Table~\ref{tab:full-compare-FS}, highlighting the effectiveness of the identity-anchored aggregator.

\begin{table}[t]
    \caption{\textbf{Generalizability evaluation.} Detectors are trained in FR and evaluated in FS to compare generalizability under the \pcolconventional setting.}
    \label{tab:generalizability}
    \centering
    \resizebox{0.95\linewidth}{!}{
        \begin{tabular}{ccccccc}
        \toprule
        MesoNet & Xception & EfficientNet & TI2Net & PWL & \model \\
        \midrule
        $0.556$ & $0.613$ & $0.738$ & $0.827$ & $0.807$ & $\textbf{0.859}$ \\
        \bottomrule
        \end{tabular}
        }
\end{table}
\paragraph{Generalizability evaluation.}
\model not only performs exceptionally in \protocolAbbrv but also demonstrates a robust ability to discern true identities from action sequences across different forgery techniques. Specifically, we evaluated a set of detectors trained under FR (using VoxCeleb) and tested them in FS (using Celeb-DF) within the \pcolconventional framework. As depicted in Table~\ref{tab:generalizability}, \model exhibits superior generalizability, outpacing all other methods in comparison. In light of this performance, we chose to employ the same \model trained in FR for all experiments, including the FS comparison (Table~\ref{tab:full-compare-FS}) and the subsequent case study (Table~\ref{tab:puppet}).

\begin{table}[t]\small
\caption{\textbf{Puppeteer re-identification results.}}
\label{tab:puppet}
\centering
\resizebox{0.95\linewidth}{!}{
    \begin{tabular}{c ccc ccc}
    \toprule
    & \multicolumn{3}{c}{FS (Celeb-DF)} & \multicolumn{3}{c}{ FR (VoxCeleb)} \\
    \cmidrule(lr){2-4} \cmidrule(lr){5-7}
    & Rank-1 & Rank-5 & mAP & Rank-1 & Rank-5 & mAP\\
    \midrule
    PWL & $91.2$ & $98.2$ & $68.0$ & $69.4$ & $89.5$ & $58.1$ \\
    A\&B & $\mathbf{99.5}$ & $\mathbf{99.7}$ & $77.0$ & $61.8$ & $87.9$ & $49.9$ \\
    ID-Reveal & $98.3$ & $99.1$ & $55.5$ & $66.2$ & $89.8$ & $54.1$ \\
    \midrule
    ID-Miner & $99.1$ & $99.4$ & $\mathbf{77.4}$ & $\mathbf{95.4}$ & $\mathbf{98.8}$ & $\mathbf{86.5}$ \\
    \bottomrule
  \end{tabular}
  }
\end{table}
\paragraph{Puppeteer re-identification (pup-reid).}
    We explore the novel task of pup-reid, which, similar to person re-id~\cite{zheng2015scalable}, aims to retrieve forgeries by the same puppeteer. Given a reference (probe), the goal is to rank all forgeries (the gallery set) based on the likelihood that the action sequence of a video originated from the same identity. Table~\ref{tab:puppet} showcases performance under FS and FR. Notably, baseline methods, which utilizes appearance-based features, find satisfactory results with FS forgeries since FS only alters the inner face region, leaving the outer face similar to that of the puppeteer. However, when faced with FR forgeries which leave minimal visual clues from the puppeteer, their performances drop significantly. In contrast, \model maintains high performance across both FS and FR since it emphasize action-based features independent of visual cues. Moreover, \model records noticeably superior results in mAP, which, unlike Rank-N, considers the quantity of true positive examples instead of only requiring one true positive within the top-N ranking. A closer look at the retrieval results reveals that baseline methods often manage to retrieve easier examples yet fail to recover the other harder ones.

\begin{figure}[t]
    \centering
    \begin{minipage}[t]{0.32\linewidth}
         \centering
         \includegraphics[width=0.78125\linewidth]{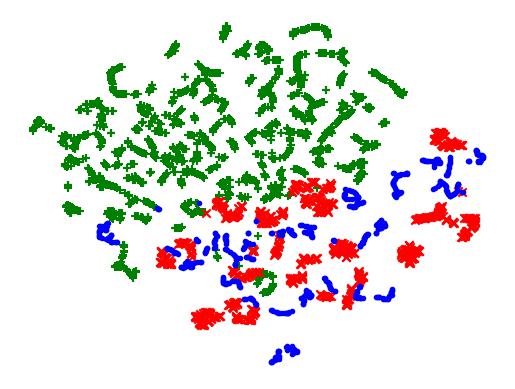}
         \pcolwhite
         \par
     \end{minipage}
     \hfill
     \begin{minipage}[t]{0.32\linewidth}
         \centering
         \includegraphics[width=0.78125\linewidth]{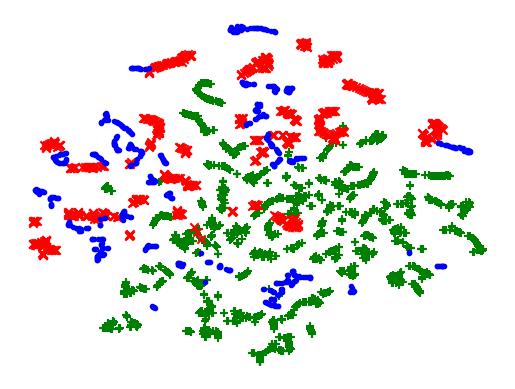}
         \pcolsur (Resize)
         \par
     \end{minipage}
     \hfill
     \begin{minipage}[t]{0.32\linewidth}
         \centering
         \includegraphics[width=0.78125\linewidth]{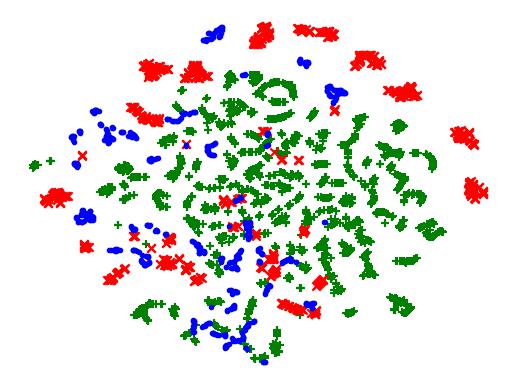}
         \pcolsur (GB)
         \par
     \end{minipage}
    \caption{\textbf{Frame distributions.} Green {\color{OliveGreen}$\mathbf{+}$}, blue {\color{blue}{$\circ$}}, red {\color{red}$\times$} represents frame sampled from $D_{gen}$, $D_{recon}$, $D_{forged}$ for \pcolwhite and $D_{gen}$, $\mathcal{A}\circ D_{gen}$, $\mathcal{A}\circ D_{forged}$ for \pcolsur (Resize and GB). The mixture of blue {\color{blue}{$\circ$}} and red {\color{red}$\times$} being separate from the green {\color{OliveGreen}$\mathbf{+}$} cluster in each plot demonstrates \protocolAbbrv reducing the distribution shift between genuine and forged examples.}
    \label{fig:tsne-rddp}
\end{figure}

\begin{figure}[t]
    \centering
    \begin{minipage}[t]{0.4\linewidth}
        \centering
        \includegraphics[width=0.625\linewidth]{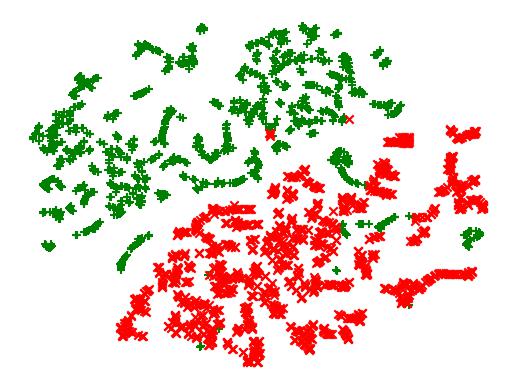}
        
        FAU attribute vectors
        \par
    \end{minipage}
    \quad\quad
    \begin{minipage}[t]{0.4\linewidth}
        \centering
        \includegraphics[width=0.625\linewidth]{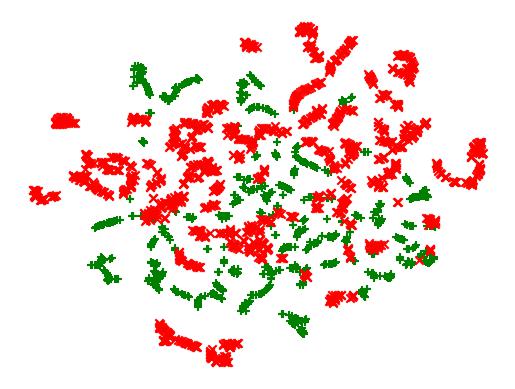}
        Artifact-agnostic encoder embeddings
        \par
    \end{minipage}
    \caption{\textbf{FAU attribute and embedding distributions.} Green {\color{OliveGreen}$\mathbf{+}$} and red {\color{red}$\times$} represents or embeddings extracted from genuine and forged examples, respectively. The contrast between clear separation (left) and mixture result (right) highlights the effectiveness of frame-level encoder in \model to be (deepfake) artifact-agnostic.}
    \label{fig:tsne-embedding}
\end{figure}

\begin{figure}[t]
    \centering
    \includegraphics[width=\linewidth]{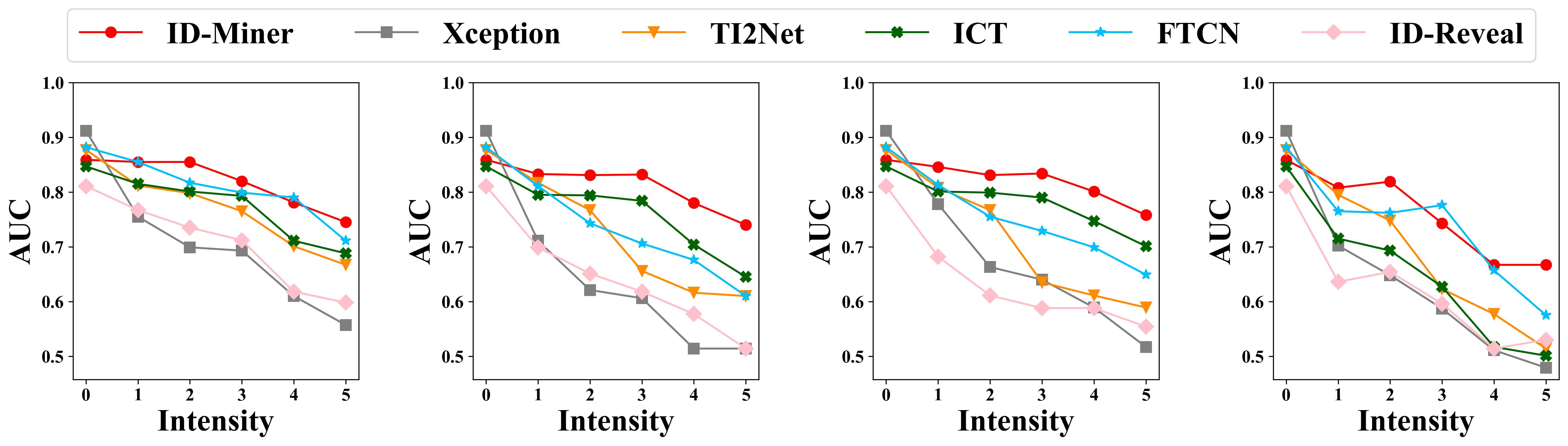}
    \begin{minipage}[t]{0.27\linewidth}\small
        \centering
        Resize
        \par
    \end{minipage}
    \hfill
    \begin{minipage}[t]{0.2\linewidth}\small
        \centering
        JPEG\\Compression
        \par
    \end{minipage}
    \hfill
    \begin{minipage}[t]{0.27\linewidth}\small
        \centering
        Video Compression
        \par
    \end{minipage}
    \hfill
    \begin{minipage}[t]{0.17\linewidth}\small
        \centering
        Gaussian Blur
        \par
    \end{minipage}
    \caption{\textbf{Sensitivity tests.} We vary the noise intensity level ranging from $0$ to $5$ for the surrogate functions under \pcolsur. \model (red) delivers comparable performance to the baseline methods under $0$ noise level while exhibiting the least degradation as noise levels increases.}
    \label{fig:sensitivity}
\end{figure}
    
\subsection{Qualitative assessments}

    We offer qualitative assessments to gain insights into the functioning of \protocolAbbrv and ID-Miner. In particular, we leverage t-SNE plots~\cite{van2008visualizing} to reveal the distribution shift under our proposed \protocolAbbrv settings, and to examine whether the frame-level encoder of \model achieves in extracting \emph{artifact-agnostic} embeddings. Subsequently, we conduct a sensitivity test across varying noise intensities under \pcolsur, providing a more in-depth exploration of \model's robustness in comparison to baseline methods.

\paragraph{Distribution alignment under \protocolAbbrv.}
    Fig.~\ref{fig:tsne-rddp} displays t-SNE plots of FAU attribute vectors of sampled frames, where the ones from genuine videos are marked by green {\color{OliveGreen}$\mathbf{+}$}, while blue {\color{blue}{$\circ$}} and red {\color{red}$\times$} denotes the corresponding \emph{compared sets} under \pcolwhite and \pcolsur.\footnote{JPEG and video compression deferred to the Appendix.} Specifically, blue {\color{blue}{$\circ$}} represents $D_{recon}$ and $\mathcal{A}\circ D_{gen}$ whereas red {\color{red}$\times$} represents $D_{forg}$ and $\mathcal{A}\circ D_{forged}$ for \pcolwhite and \pcolsur, respectively. As shown in all three plots, the clustering of green {\color{OliveGreen}$\mathbf{+}$} points and the tendency of blue {\color{blue}{$\circ$}} and red {\color{red}$\times$} points to mix together indicate that \protocolAbbrv successfully aligns the distribution between forged and genuine examples. Furthermore, the separation of green and red samples under \pcolwhite provides evidence that a significant distribution shift exists between $D_{gen}$ and $D_{forg}$.

\paragraph{Effectiveness of artifact-agnostic encoder.}
    Figure~\ref{fig:tsne-embedding} provides a comparative visualization of t-SNE plots for FAU attribute vectors and frame-level encoder embeddings of \model. 
    In both plots, green {\color{OliveGreen}$\mathbf{+}$} represents attribute vectors or embeddings derived from genuine video frames while red {\color{red}$\times$} represents those from forged videos. Notice the clear partition between genuine (green) and forged (red) FAU vectors, compared with the amalgamated mixture for the encoder embeddings. Such contrast demonstrates the effectiveness of \emph{artifact-agnostic loss}, which guides the frame-level encoder of \model to produce the same embedding features for the same facial `pose', irrespective of artifacts or appearance change caused by the deepfake processes.

\paragraph{Sensitivity tests.}
    We conduct sensitivity tests using \pcolsur under different noise intensity levels from $0$ to $5$.\footnote{A level of $0$ indicates no perturbation (\pcolconventional); we set $3$ as default for all other experiments.} As depicted in Fig.~\ref{fig:sensitivity}, \model (red line) demonstrates robust performances across all noise levels, as it generally outperforms all baseline methods  under noise. This resilience underscores \model's effectiveness in mining out the identity despite the added perturbation, highlighting its potential for practical detection applications in real-world scenarios.
\section{Discussions and Conclusion}

Recently, several researchers expressed concerns about the rapid development of generative AI, fearing a world where authenticity and truth become elusive. Indeed, while deepfakes grow increasingly sophisticated, there is an escalating need for advanced detection methods, yet progress in detection often lags behind the pace of deepfake. In this work, we introduce a proactive approach to the detection race, preemptively countering ``perfect deepfakes.'' Our novel \protocol (\protocolAbbrv) effectively aligns the distributions of forged and genuine examples using white-hat deepfake algorithms (\pcolwhite) or surrogate functions (\pcolsur). The significant disparity in baseline detection performances between \pcolconventional and \protocolAbbrv highlights the limitations of existing methods that rely on artifact-induced distribution shifts. In response, we propose \model, a novel detection model that ignores deepfake-induced artifacts and appearance variations. By incorporating the \emph{identity-anchored loss} and the \emph{artifact-agnostic loss}, \model excels under the challenging evaluations of \protocolAbbrv. Summarily, our work represents an initial step toward detecting ``perfect deepfakes.'' Although \model provides an approach to action sequence-based identification, we advocate for future works to explore deeper analysis into human pose and motion behaviours for the verification of identities portrayed in a video. However, we firmly advise against extrapolation of the same principle to create more intricate deepfakes. Specifically, attempts to \textit{mimic} genuine or habitual actions of individuals infringe upon their right to identity; adherence to ethical guidelines is urged.

\bibliography{ref}

\clearpage
\appendix
\section{Broader Impact}
Our work in deepfake detection carries implications beyond the immediate boundaries of our research. We outline both the positive and negative implications, shedding light on the potential societal ramifications of our discoveries.

\paragraph{Positive impact.} 
The innovation of both \emph{\protocol (\protocolAbbrv)} and \emph{Identity-anchored Artifact-agnostic Deepfake Detection (\model)} contributes to the strengthening of deepfake detection techniques. As deepfake technology evolves and generates increasingly imperceptible artifacts, leading to outputs that are progressively harder to differentiate from genuine instances, our research strengthens the defensive measures against these sophisticated forgeries. \protocolAbbrv reveals the dependency of existing detection methods on simplistic artifact features. Meanwhile, \model pioneers a new detection approach to be \emph{artifact-agnostic}, contrasting various facial actions irrespective of deepfake-related artifacts. Moreover, it promotes \emph{identity-anchored} detection by contrasting deepfake processed samples to extract distinguishing action sequence features, irrespective of artifact and appearance. Despite the recent development of generative AI, these progressions can reinstate confidence in digital communication. Furthermore, we are confident that our research sets the stage for more comprehensive investigations into proactively addressing potential AI risks, fostering innovation, and propelling advancements to counteract the dangers posed by deepfakes. Specifically, our work encourages future studies to delve deeper into utilizing action sequences to identify the individual responsible for performing these actions.

\textbf{Negative impact.} 
However, our efforts might unintentionally fuel the competition between deepfake creators and deepfake detection systems. When we expose the existing flaws in detection methods, malicious individuals may exploit this information to create even more advanced forgeries, thereby increasing the difficulty of deepfake detection. Furthermore, there are concerns regarding the potential misuse of our framework. Authoritarian governments might alter it to engage in surveillance and manipulation tactics. At the same time, its widespread application could unintentionally violate people's privacy by revealing more personal information in videos than initially intended. Therefore, it is crucial to emphasize the significance of related research methodologies and adherence to ethical guidelines to minimize potentially unfavorable outcomes. Researchers must carefully contemplate the societal ramifications of their work and work towards developing solutions that prioritize the safety and welfare of individuals and communities.

\section*{Limitations}
We address the limitations of this work. Firstly, although \protocolAbbrv serves as an initial solution for balancing the distribution between genuine and forged examples, its robustness against highly sophisticated deepfakes still needs to be tested. A more rigorous theory addressing distribution shift needs to be developed. Also, future research should stress test our proposed \model against emerging deepfake techniques. Secondly, \model is designed primarily for the portrait video format. Thus, it may be challenging to apply \model to deepfakes involving full-person or multi-person videos, where facial regions constitute a smaller proportion of the frame. Moreover, \model's emphasis on action sequences presents challenges for detecting deepfakes within single frames or still images. Thirdly, \model assumes the availability of a reference video or some knowledge about the person being examined. Misjudgments in identifying the targeted individual---potentially due to adversarial attacks against face recognition systems---may lead to incorrect results. Finally, our frameworks depend on the availability of sufficient training data. In situations where data are scarce, especially for individuals or specific contexts that are less frequently portrayed, such as explicit content, the performance of our methods may be compromised. These limitations present opportunities for future research, highlighting the necessity for ongoing progress and adjustment in response to the evolution of deepfake algorithms.
\section{Additional implementation details}
\label{app:info}
In the following, we present the implementation details for \model in Section~\ref{subA:training}, testing procedures under conventional and \protocolAbbrv in Section~\ref{subA:testing}. Besides, details of the dataset and baselines are also provided in Section~\ref{subA:baseline} and Section~\ref{subA:baseline}, respectively. For reader clarity, we include a table of notations in Table~\ref{tab:notation} and a table of abbreviations in Table~\ref{tab:abbreviation}.\footnote{We provide the anonymized repository link to our source code for review: \url{https://anonymous.4open.science/r/idminer-7F15/}}

\begin{table*}[t]
    \caption{\textbf{Notation table.}}
    \label{tab:notation}
    \centering
    \begin{tabular}{c p{.8\textwidth}}
    \toprule
    symbol & description \\
    \midrule
    $\mathbf{a}, \mathbf{a}', \mathbf{b}$ & An action sequence. We use $\mathbf{a}$ and $\mathbf{a}'$ to indicates action sequences of the same person while $\mathbf{b}$ represent that of a different identity. \\
    $s,t,x,y,z$ & Different identities. When used as deepfake inputs, \eg, $\mathcal{F}_s$, the notation indicates that their portrait images are utilized for deepfake algorithm to create forgeries with their appearance.\\
    $V(s,\mathbf{a})$ & A video with subject (appearance) $s$ and action sequence $\mathbf{a}$; subscript $i$ denotes the $i_{th}$ frame as $V_i$.\\ 
    $\mathcal{F}_s$ & A deepfake function that produce forgery with the appearance of subject $s$.\\
    $\mathcal{A}$ & The surrogate function to apply noise to both forged and genuine examples sets.\\
    $D$ & A set of examples pertaining to a real or fake class; $D_{gen}$, $D_{forg}$, $D_{recon}$, $\tilde{D}_{gen}$, $\tilde{D}_{forg}$ represents the genuine, forged, reconstructed, approximated genuine, and approximated forged sets, respectively.\\
    $\mathcal{L}_{artifact}$ & The artifact-agnostic loss function. \\
    $\mathcal{L}_{identity}$ & The identity-anchored loss function. \\
    $\mathcal{L}_{total}$ & The total training loss $\mathcal{L}_{total} = \mathcal{L}_{identity} + \lambda\mathcal{L}_{artifact}$. \\
    $\tau$, $\lambda$ & The temperature parameter for contrastive losses~\cite{chen2020simple} and the hyperparameter to balance between the \emph{artifact-agnostic loss} and the \emph{identity-anchored loss}.\\
    $\mathcal{E}$ & The frame-level encoder that outputs the frame-level embeddings.\\
    $\mathbf{q_i}$, $\mathbf{k^+_i}$, $\mathbf{k^-_j}$ & The frame-level embeddings selected as the query, positive, and negative samples in the \emph{artifact-agnostic loss}. \\
    $\mathcal{M}$ & The entire \model that outputs the video representations.\\
    $\mathbf{q}$, $\mathbf{k^+}$, $\mathbf{k^-}$ & The video-level representation selected as the query, positive, and negative samples in the \emph{identity-anchored loss}.\\
    $(\_\cdot\_)$ & Cosine similarity between two vectors. \\
    \bottomrule
    \end{tabular}
\end{table*}

\begin{table*}[t]
    \caption{\textbf{Abbreviation table.}}
    \label{tab:abbreviation}
    \centering
    \begin{tabular}{c p{.8\textwidth}}
    \toprule
    \protocolAbbrv & \protocol, a novel evaluation setting aimed at reducing the distribution shift between genuine and forged examples. In contrast, \pcolconventional denotes the setting where genuine and forged examples are directly used, allowing detectors to rely on the deepfake-induced artifacts.\\
    \multirow{2}{*}{\shortstack[l]{RDDP-\\\textsc
        {whitehat}}} & The first variant of \protocolAbbrv, where a white-hat deepfake algorithm is employed to reconstruct the genuine examples each using a portrait of the same subject. As such, the resulting reconstructed examples exhibits the deepfake artifacts yet have the same appearance and action sequences.\\
    \multirow{2}{*}{\shortstack[l]{RDDP-\\\textsc
        {surrogate}}}  & The second variant of \protocolAbbrv, where surrogate functions are utilized to introduce universal noise to both forged and genuine examples. The surrogates in this work include resize, JPEG compression, video compression, and Gaussian noise\\
    \model & Our proposed detection model featuring \emph{artifact-agnostic loss} at the frame level and \emph{identity-anchored loss}  at the video level.\\
    \bottomrule
    \end{tabular}
\end{table*}

\subsection{\model training detail}
\label{subA:training}

    Before the training stage, we utilize the First Order Motion Model (FOMM)~\cite{siarohin2019first} to augment our training dataset, referred to as $D_{gen}$, which initially consists of genuine videos only. This augmentation process results in an augmented dataset denoted as $D_{aug}$. The FOMM model employs self-supervised learning to learn about the local affine transformations at the detected key points. Furthermore, the FOMM model can transfer the facial motion from the video onto the source image by providing a driving video and a source image. We chose this model due to its availability to the public, satisfactory quality, computational efficiency, and capability to generate large-scale forged videos using a single model. During the training phase, with a batch size of $64$, we randomly select $8$ classes from the augmented dataset $D_{aug}$, ensuring that each class contributes $8$ videos. Additionally, we retrieve the corresponding original driving videos from the original dataset $D_{gen}$. Subsequently, we proceed to identify positive and negative pairs within the batch.
    
    In addition, we designate the original driving video from $D_{gen}$ as the query for frame-level contrastive learning. The corresponding forged videos from $D_{aug}$ serve as the positive examples, while the negative examples are exhaustively chosen from different classes within the batch from $D_{aug}$. Finally, we apply the loss function on a per-frame basis. Moreover, the query consists of videos from $D_{aug}$ for video-level contrastive learning. The positive examples are videos with the same identity as the query but different appearances, also obtained from $D_{aug}$. The negative examples are videos from different classes.
    
\subsection{Conventional and \protocolAbbrv testing procedures}
\label{subA:testing}

    Our experiments are based on large-scale portrait video datasets, including VoxCeleb~\cite{nagrani2017voxceleb} with over $20$k videos across $1251$ subjects, and Celeb-DF~\cite{li2020celeb} consisting of $590$ genuine videos, $5639$ deepfake videos over $59$ subjects. We establish two divisions of detection evaluations corresponding to \textit{face reenactment (FR)} and \textit{faceswap (FS)}. Note that the identities in the testing set are disjoint from our training set, and we maintain the same set of videos across three protocols.

    For FR, the original VoxCeleb dataset serves as $D_{gen}$, while forged videos are generated as $D_{forg}$ using the FOMM model. Additionally, $D_{recon}$ is formed by randomly selecting a frame from a different video with the same identity as the source image. For FS, $D_{gen}$ and $D_{forg}$ are derived from the original dataset. Subsequently, we generate $D_{recon}$ by MobileFaceSwap\cite{xu2022MobileFaceSwap}, where a frame is randomly selected from a different video that shares the same identity as the source image.

    \textbf{Reference-free detectors:} For the models that produce an output from a single input, we denote each testing video as $V$ and the ground truth as $y$, we clarify the video type and label under different protocols:
    
    \begin{equation}
    \label{eqn:no_reference_conventional}
        \pcolconventional:
        \begin{cases}
            V\in D_{gen}, & y=1 \\
            V\in D_{forg}, & y=0
        \end{cases},
    \end{equation}

    \begin{equation}
    \label{eqn:no_reference_rddpwh}
        \pcolwhite:
        \begin{cases}
            V\in D_{recon}, & y=1 \\
            V\in D_{forg}, & y=0
        \end{cases},
    \end{equation}

    \begin{equation}
    \label{eqn:no_reference_rddpsur}
        \pcolsur:
        \begin{cases}
            V\in A\circ D_{gen}, & y=1 \\
            V\in A\circ D_{forg}, & y=0
        \end{cases}.
    \end{equation}

    \textbf{Reference-based detectors:} In the case of models that necessitate a reference video, we ensure the availability of a video depicting the same identity as the examined video. We denote each testing video as $V$, its reference video as $R$, and the ground truth as $y$, we clarify the video type and label under different protocols:

    \begin{equation}
    \label{eqn:reference_conventional}
        \fontsize{6.5pt}{6.5pt}{\pcolconventional}:
        \begin{cases}
            V\in D_{gen} ,\: R\in D_{gen}, & y=1 \\
            V\in D_{forg} ,\: R\in D_{gen}, & y=0 
        \end{cases},
    \end{equation}
    
    \begin{equation}
    \label{eqn:reference_rddpwh}
        \fontsize{6.5pt}{6.5pt}{\pcolwhite}:
        \begin{cases}
            V\in D_{gen} ,\: R\in D_{gen}, & y=1 \\
            V\in D_{forg} ,\: R\in D_{recon}, & y=0 
        \end{cases},
    \end{equation}

    \begin{equation}
    \label{eqn:reference_rddpsur}
        \fontsize{6.5pt}{6.5pt}{\pcolsur}:
        \begin{cases}
            V\in D_{gen} ,\: R\in D_{gen}, & y=1 \\
            V\in A\circ D_{forg} ,\: R\in A\circ D_{gen}, & y=0 
        \end{cases}.
    \end{equation}

    The evaluation process for all detectors in three different protocols adheres to a consistent rule, where each model produces a "score" ranging from 0 to 1, which determines the authenticity of a sample. The specific nature of these scores varies depending on the design of each detector. They may be in the form of logits, representing the probability of a video being genuine, or similarity scores indicating the resemblance between the sample and a reference. In cases where the model outputs embedding distances, we compute the reciprocal of the distance to derive the similarity score. All metrics are reported at the video-level. If the model operates on a per-frame basis, we calculate the average output across all frames to obtain the final result. Finally, the scores for the entire dataset are collected, and an algorithm is applied to calculate the Area Under the Curve (AUC).
    
    It is important to note that the detection methods used in FWA, Face X-ray, EFNB4 +SBIs, and ICT assume that the frames have blended boundaries between the manipulated region and the genuine part. Therefore, we did not report the AUC in Table \ref{tab:full-compare-FR} since the frames in these cases are fully synthetic, which means all the pixels are generated by the model.

\subsection{Baseline details}
\label{subA:baseline}
    We introduce each of the compared baselines as follows.
    \begin{itemize}
        \item \textbf{Xception}~\cite{chollet2017xception} and \textbf{EfficientNet}~\cite{tan2019efficientnet}. Although these methods are not specifically designed for deepfake detection, they are often used as baselines due to their performances.
        \item \textbf{MesoNet}~\cite{afchar2018mesonet} is a deep neural network with a small number of layers. This approach is placed at a mesoscopic level of analysis, which is an intermediate approach between microscopic and semantic levels.
        \item \textbf{FWA}~\cite{li2018exposing} is based on the observation that current deepFake algorithms can only generate images of limited resolutions, which need to be further warped to match the original faces in the source video.
        \item \textbf{Face-X-ray}~\cite{li2020face} is based on the observation that most existing face manipulation methods share a common blending step, and there exist intrinsic image discrepancies across the blending boundary, which is neglected in advanced face manipulation detectors.
        \item \textbf{FTCN}~\cite{zheng2021exploring} consists of two major stages. The first stage is a fully temporal convolution network (FTCN) that reduces the spatial convolution kernel size to 1 while maintaining the temporal convolution kernel size unchanged. This design benefits the model for extracting temporal features and improves generalization capability. The second stage is a Temporal Transformer network that explores long-term temporal coherence.
        \item \textbf{EFNB4+SBIs}~\cite{shiohara2022detecting} use synthetic training data called self-blended images (SBIs), which are generated by blending pseudo source and target images from single genuine images, reproducing common forgery artifacts.
        \item \textbf{ICT}~\cite{dong2022protecting} is based on the observation that the inner face and outer face are inconsistent in faceswap forgeries.
        \item \textbf{TI2Net}~\cite{liu2023ti2net} is a reference-agnostic detector focusing on temporal identity inconsistency, i.e., the low similarity of identity features captured from the same video with the given identity.
        \item \textbf{PWL}~\cite{agarwal2019protecting} is an identity-specific model that computes the correlation of facial action units in videos associated with a specific identity. It then employs an one-class SVM to identify outliers.
        \item \textbf{A\&B}~\cite{agarwal2020detecting} is the pioneering work in deepfake detection that leverages reference videos as guidance to verify the examined video based on both its appearance (A) and behavior (B).
        \item \textbf{ID-Reveal}~\cite{idreveal2021cozzolino} is an identity-aware approach that utilizes an adversarial training strategy to guide the encoder in learning identity-aware motion.
    \end{itemize}
    Note that due to the variety in training approaches adopted by baseline methods and the absence of publicly released code for some, not all baseline preparations are executed under identical setups. Nevertheless, for each baseline, our aim is to prepare separate versions for Face Recognition (FR) and Face Swap (FS) testing, specifically using VoxCeleb for FR and CelebDF for FS. Moreover, we ensure that the identities in the training set do not overlap with those in the testing set when preparing identity-based detections. For methods that process single frames~\cite{chollet2017xception,afchar2018mesonet,tan2019efficientnet,li2018exposing, shiohara2022detecting, dong2022protecting}, we compute the detection result as the average across all frames. For~\cite{idreveal2021cozzolino,zheng2021exploring,li2018exposing,shiohara2022detecting, dong2022protecting}, we employ the pre-trained weights made available by the authors, given the absence of publicly released training code. It is worth mentioning that for the method proposed in \cite{agarwal2019protecting}, we adhere to the procedure outlined in their work and prepare a distinct model for each identity in the testing set. As for our proposed detection approach, \model, we capitalize on its inherent generalizability. Specifically, \model is solely trained on the Face Recognition (FR) division using the VoxCeleb dataset and its corresponding deepfake augmentations.

\section{Additional evaluations and visualizations}
\label{app:exp}

\subsection{Additional quantitative results}

\subsubsection{Evaluations under other metrics.}
    Following~\cite{guan2022delving, chen2022ost, idreveal2021cozzolino}, we present the accuracy (ACC) results in Table~\ref{apptab:FR-ACC} and Table~\ref{apptab:FS-ACC} for the same experimental evaluations as shown in Table~\ref{tab:full-compare-FR} and Table~\ref{tab:full-compare-FS}, which display AUC values. Similar to the findings in Table~\ref{tab:full-compare-FR} and Table~\ref{tab:full-compare-FS} which provide AUC measurements, we observe a decline in performance where most of the baseline methods demonstrate excellent results under \pcolconventional with $0.8$ to $0.9$ accuracy, yet decrease dramatically under \protocolAbbrv, with an average drop ranging from $7\%$ to $32\%$. On the other hand, our \model only experience $4\%$ and $4.6\%$ drop in FR and FS, respectively. 
    
    \begin{table*}[t]\small
    \caption{\textbf{Detection evaluation for face reenactment (FR) in ACC.}}
    \label{apptab:FR-ACC}
    \centering
    \resizebox{0.95\textwidth}{!}{
    \begin{tabular}{l c c cccc c}
        \toprule
        & \multirow{3}{*}{\shortstack[l]{\textsc{conven}\\ -\textsc{tional}}} 
        & \multirow{3}{*}{\shortstack[l]{RDDP-\\\textsc
        {whitehat}}} 
        & \multicolumn{4}{c}{\pcolsur} 
        & \multirow{2}{*}{\shortstack[l]{\\ \textit{avg. drop}}} \\
        \cmidrule(lr){4-7} 
        &                   & & Resize & JPEG &  Video Compression & Gaussian Blur \\
        \midrule
        Xception~\cite{chollet2017xception} & $\underline{0.855}$ &$ 0.525$ & $0.595$ & $0.510$ & $0.535$ & $0.495$ & $-32.2\%$ \\
        MesoNet~\cite{afchar2018mesonet} & 0$.830$ & $0.555$ & $0.595$ & $0.515$ & $0.515$ & $0.495$ & $-29.5\%$\\
        EfficientNet~\cite{tan2019efficientnet} & $\mathbf{0.870}$ & $0.670$ & $0.670$ & $0.690$ & $0.685$ & $0.645$ & $-19.8\%$ \\
        FTCN~\cite{zheng2021exploring} & $0.840$ & $0.595$ & $0.665$ & $0.665$ & $0.625$ & $0.635$ & $-20.3\%$ \\
        TI2Net~\cite{liu2023ti2net} & $0.850$ & $0.615$ & $\underline{0.725}$ & $0.640$ & $0.625$ & $0.625$ & $-20.4\%$ \\
        \midrule
        PWL~\cite{agarwal2019protecting} &$ 0.780$ & $0.630$ & $0.635$ & $0.630$ & $0.645$ & $0.635$ & $-14.5\%$ \\
        A\&B~\cite{agarwal2020detecting} & $0.575$ & $0.560$ & $0.540$ & $0.540$ & 0.565 & $0.565$ & $\mathbf{-2.1\%}$ \\
        ID-Reveal~\cite{idreveal2021cozzolino} & $0.680$ & $0.530$ & $0.640$ & 0.560 & $0.540$ & $0.515$ & $-12.3\%$ \\
        \midrule
        \model & $0.780$ & $\mathbf{0.750}$ & $\mathbf{0.750}$ & $\mathbf{0.760}$ & $\mathbf{0.760}$ & $\mathbf{0.680}$ & \underline{$-4\%$} \\
        \model (no FLE) & $0.835$ & $\underline{0.710}$ & $0.715$ & $\underline{0.755}$ & $\underline{0.730}$ & $\underline{0.665}$ & $-12.0\%$ \\
        \bottomrule
    \end{tabular}
    }
\end{table*}

\begin{table*}[t]\small
    \caption{\textbf{Detection evaluation for faceswap (FS) in ACC.}}
    \label{apptab:FS-ACC}
    \centering
    \resizebox{0.95\textwidth}{!}{
    \begin{tabular}{l c c cccc c}
        \toprule
        & \multirow{3}{*}{\shortstack[l]{\textsc{conven}\\ -\textsc{tional}}} 
        & \multirow{3}{*}{\shortstack[l]{RDDP-\\\textsc
        {whitehat}}} 
        & \multicolumn{4}{c}{\pcolsur} 
        & \multirow{2}{*}{\shortstack[l]{\\ \textit{avg. drop}}} \\
        \cmidrule(lr){4-7} 
        &                   & & Resize & JPEG &  Video Compression & Gaussian Blur \\
        \midrule
        Xception~\cite{chollet2017xception} & $\mathbf{0.835}$ & $0.595$ & $0.625$ & $0.545$ & $0.585$ & $0.525$ & $-26.0\%$\\
        MesoNet~\cite{afchar2018mesonet} & $0.690$ & $0.515$ & $0.485$ & $0.460$ & $0.465$ & $0.475$ & $-21.0\%$\\
        FWA~\cite{li2018exposing} & $0.605$ & $0.535$ & $0.550$ & $0.535$ & $0.535$ & $0.510$ & $-7.2\%$\\
        EfficientNet~\cite{tan2019efficientnet} & $0.800$ & $0.620$ & $0.640$ & $0.620$ & $0.610$ & $0.605$ & $-18.1\%$\\
        Face X-ray~\cite{li2020face} & $0.790$ & $0.600$ & $0.615$ & $0.585$ & $0.565$ & $0.535$ & $-21.0\%$\\
        FTCN~\cite{zheng2021exploring} & $\underline{0.825}$ & $0.600$ & $0.705$ & $0.635$ & $0.640$ & $\mathbf{0.685}$ & $-17.2\%$\\
        EFNB4+SBIs~\cite{shiohara2022detecting} & $0.800$ & $0.595$ & $0.660$ & $0.635$ & $0.590$ & $0.525$ & $-20.0\%$\\
        ICT\cite{dong2022protecting} & $0.755$ & $0.580$ & $\underline{0.720}$ & $0.710$ & $\underline{0.715}$ & $0.585$ & $-9.3\%$\\
        TI2Net~\cite{liu2023ti2net} & $0.785$ & $0.620$ & $0.690$ & $0.630$ &$ 0.625$ & $0.625$ & $-14.7\%$\\
        \midrule
        PWL~\cite{agarwal2019protecting} & $0.800$ & $0.600$ & $0.600$ & $0.600$ & $0.615$ & $0.615$ & $-19.4\%$\\
        A\&B~\cite{agarwal2020detecting} & $0.530$ & $0.485$ & $0.475$ & $0.475$ & $0.500$ & $0.495$ & $\underline{-4.4\%}$\\
        ID-Reveal~\cite{idreveal2021cozzolino} & $0.750$ & $0.575$ & $0.670$ & $0.600$ & $0.590$ & $0.590$ & $-14.5\%$\\
        \midrule
        \model & $0.770$ & $\mathbf{0.735}$ & $\mathbf{0.735}$ & $\mathbf{0.750}$ & $\mathbf{0.750}$ & $\underline{0.650}$ & $\mathbf{-4.6\%}$\\
        \model (no FLE) & $0.795$ & $\underline{0.675}$ & $0.685$ & $\underline{0.715}$ & $0.700$ & $0.615$ & $-11.7\%$\\
        \bottomrule
    \end{tabular}
    }
\end{table*}

\begin{table*}[t]\small
    \caption{\textbf{Evaluations of baseline methods trained under RDDP (FR).}}
    \label{tab:train-on-RDDP-FR}
    \centering
    \resizebox{0.8\textwidth}{!}{
    \begin{tabular}{l c cccc}
        \toprule
        & \multirow{3}{*}{\shortstack[l]{RDDP-\\\textsc
        {whitehat}}} 
        & \multicolumn{4}{c}{\pcolsur} \\
        \cmidrule(lr){3-6} 
        &                   & Resize & JPEG & Video Compression & Gaussian Blur \\
        \midrule
        Xception~\cite{chollet2017xception} & $0.605$ & $0.681$ & $0.599$ & $0.635$ & $0.588$ \\
        MesoNet~\cite{afchar2018mesonet} & $0.655$ & $0.670$ & $0.575$ & $0.599$ & $0.557$ \\
        EfficientNet~\cite{tan2019efficientnet}  & $0.769$ & $0.758$ & $0.795$ & $0.771$ & $0.741$  \\
        TI2Net~\cite{liu2023ti2net} & $0.689$ & $0.808$ & $0.705$ & $0.699$ & $0.661$ \\
        \midrule
        PWL~\cite{agarwal2019protecting} & $0.679$ & $0.701$ & $0.715$ & $0.699$ & $0.681$ \\
        A\&B~\cite{agarwal2020detecting} & $0.610$ & $0.615$ & $0.596$ & $0.620$ & $0.625$ \\
        \bottomrule
    \end{tabular}
    }
\end{table*}

\begin{table*}[t]\small
    \caption{\textbf{Evaluations of baseline methods trained under RDDP (FS).}}
    \label{tab:train-on-RDDP-FS}
    \centering
    \resizebox{0.8\textwidth}{!}{
    \begin{tabular}{l c cccc}
        \toprule
        & \multirow{3}{*}{\shortstack[l]{RDDP-\\\textsc
        {whitehat}}} 
        & \multicolumn{4}{c}{\pcolsur} \\
        \cmidrule(lr){3-6} 
        &                   & Resize & JPEG & Video Compression & Gaussian Blur \\
        \midrule
        Xception~\cite{chollet2017xception} & $0.667$ & $0.693$ & $0.615$ & $0.655$ & $0.599$ \\
        MesoNet~\cite{afchar2018mesonet} & $0.554$ & $0.511$ & $0.520$ & $0.517$ & $0.538$ \\
        EfficientNet~\cite{tan2019efficientnet}  & $0.719$ & $0.739$ & $0.686$ & $0.692$ & $0.647$  \\
        TI2Net~\cite{liu2023ti2net} & $0.622$ & $0.795$ & $0.667$ & $0.671$ & $0.659$ \\
        \midrule
        PWL~\cite{agarwal2019protecting} & $0.668$ & $0.712$ & $0.735$ & $0.709$ & $0.699$ \\
        A\&B~\cite{agarwal2020detecting} & $0.585$ & $0.627$ & $0.554$ & $0.571$ & $0.587$ \\
        \bottomrule
    \end{tabular}
    }
\end{table*}

\subsubsection{Training baseline methods under RDDP.}
We present extended results for baseline methods trained on FR and FS datasets modified per \pcolwhite guidelines in Table~\ref{tab:train-on-RDDP-FR} and Table~\ref{tab:train-on-RDDP-FS}. Specifically, we subject these baseline models to the same training regimen as \model to discern if their subpar performance is attributed to unfamiliarity with the demanding \protocolAbbrv setting. When juxtaposed with the results from Table~\ref{tab:full-compare-FR} and Table~\ref{tab:full-compare-FS}, it becomes evident that baseline methods continue to underperform even after training within the \protocolAbbrv environment. This is because their methods lean heavily on the distributional disparities between genuine and deepfake videos. Conversely, \model is designed to overlook artifacts, emphasizing the extraction of robust identity features rooted in the action sequences of a portrait video. As a result, while baseline methods falter under \protocolAbbrv---irrespective of whether trained conventionally or under \protocolAbbrv---\model consistently surpasses them in \protocolAbbrv evaluations.

\subsection{Additional qualitative results}
\subsubsection{Distribution alignment under \protocolAbbrv.}

\begin{figure*}[t]
    \centering
    \begin{minipage}[t]{0.45\textwidth}
         \centering
         \includegraphics[width=\textwidth]{fig/tsne_rddpwh.jpg}
         \pcolwhite
         \par
     \end{minipage}
     \hfill
     \begin{minipage}[t]{0.45\textwidth}
         \centering
         \includegraphics[width=\textwidth]{fig/tsne_rddprz.jpg}
         \pcolsur (Resize)
         \par
     \end{minipage}
     \hfill
     \begin{minipage}[t]{0.45\textwidth}
         \centering
         \includegraphics[width=\textwidth]{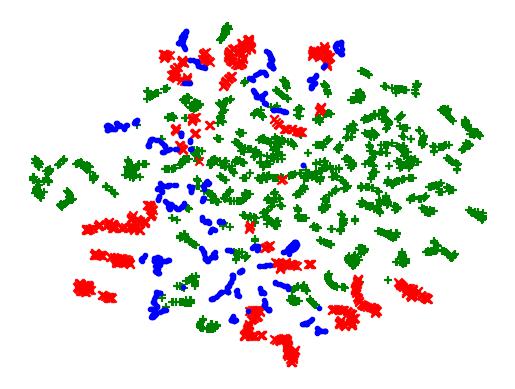}
         \pcolsur (JPEG compression)
         \par
     \end{minipage}
     \hfill
     \begin{minipage}[t]{0.45\textwidth}
         \centering
         \includegraphics[width=\textwidth]{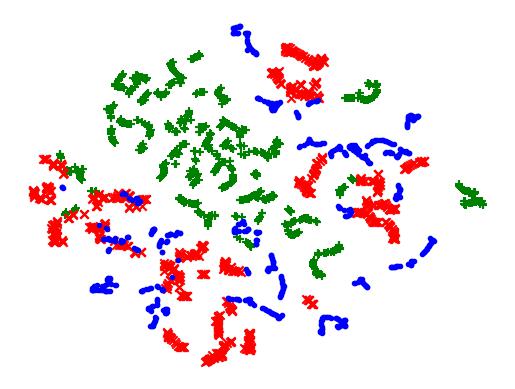}
         \pcolsur (Video compression)
         \par
     \end{minipage}
     \hfill
     \begin{minipage}[t]{0.45\textwidth}
         \centering
         \includegraphics[width=\textwidth]{fig/tsne_rddpgb.jpg}
         \pcolsur (Gaussian blur)
         \par
     \end{minipage}
    \caption{\textbf{Frame distributions.} We provide t-SNE plots of FAU attribute vectors of sampled testing frames under the \protocolAbbrv. Green {\color{OliveGreen}$\mathbf{+}$}, blue {\color{blue}{$\circ$}}, red {\color{red}$\times$} each represents frame sampled from $D_{gen}$, $D_{recon}$, $D_{forged}$ for \pcolwhite and $D_{gen}$, $\mathcal{A}\circ D_{gen}$, $\mathcal{A}\circ D_{forged}$ for \pcolsur (Resize, JPEG compression, Video compression and Gaussian blur). The mixture of blue {\color{blue}{$\circ$}} and red {\color{red}$\times$} being separate from the green {\color{OliveGreen}$\mathbf{+}$} cluster in each plot demonstrates \protocolAbbrv reducing the distribution shift between genuine and forged examples.}
    \label{appfig:all-tsne-rddp}
\end{figure*}

In Fig.~\ref{appfig:all-tsne-rddp}, we extend our examination of frame distribution to include t-SNE plots for both \pcolwhite and all four surrogate functions under \pcolsur. This includes JPEG compression and video compression under \pcolsur, which due to space limitations, were absent from the main body of the paper. As depicted in Fig.~\ref{appfig:all-tsne-rddp}, applying JPEG compression and video compression under \pcolsur efficiently bridges the distribution disparity between genuine and forged instances. These additional plots, mirroring the established pattern seen in other plots in the main paper, show a mix of blue {\color{blue}{$\circ$}} (representing $\tilde{D}_{gen}$) and red {\color{red}$\times$} (representing $\tilde{D}_{forg}$) samples that are distinctly separated from the green {\color{OliveGreen}$\mathbf{+}$} cluster. These added visualizations further corroborate the efficacy of \protocolAbbrv in reducing the distribution shift between forged and genuine examples.

\subsection{Data sample visualizations}

    \paragraph{Training environment of \model.}

\begin{figure*}[t]
    \centering
    
     \begin{minipage}[t]{0.19\textwidth}
         \centering
         \phantom{\rule{0pt}{\textwidth}} 
         \par
     \end{minipage}
     \begin{minipage}[t]{0.19\textwidth}
         \centering
         \includegraphics[width=\textwidth]{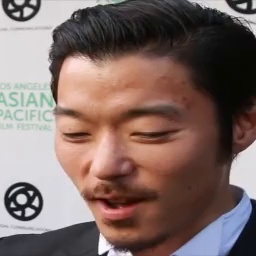}
         \par
     \end{minipage}
     \begin{minipage}[t]{0.19\textwidth}
         \centering
         \includegraphics[width=\textwidth]{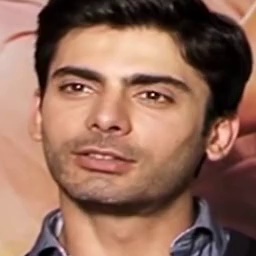}
         \par
     \end{minipage}
     \begin{minipage}[t]{0.19\textwidth}
         \centering
         \includegraphics[width=\textwidth]{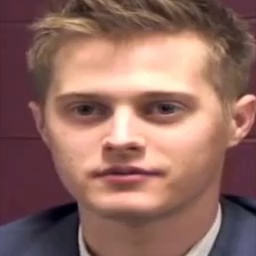}
         \par
     \end{minipage}
     \begin{minipage}[t]{0.19\textwidth}
         \centering
         \includegraphics[width=\textwidth]{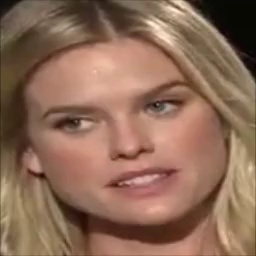}
         \par
     \end{minipage}
     \begin{minipage}[t]{0.19\textwidth}
         \centering
         \includegraphics[width=\textwidth]{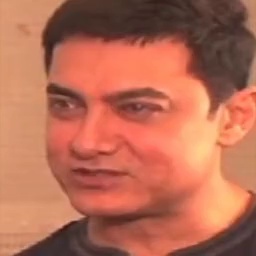}
         \par
     \end{minipage}
     \begin{minipage}[t]{0.19\textwidth}
         \centering
         \includegraphics[width=\textwidth]{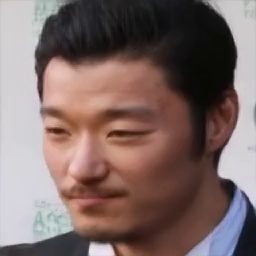}
         \par
     \end{minipage}
     \begin{minipage}[t]{0.19\textwidth}
         \centering
         \includegraphics[width=\textwidth]{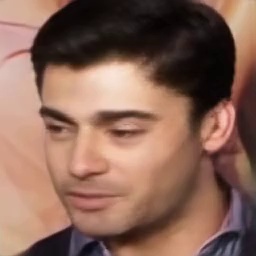}
         \par
     \end{minipage}
     \begin{minipage}[t]{0.19\textwidth}
         \centering
         \includegraphics[width=\textwidth]{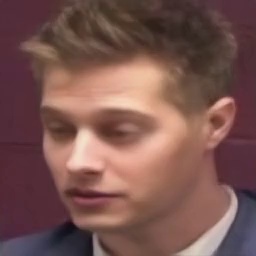}
         \par
     \end{minipage}
     \begin{minipage}[t]{0.19\textwidth}
         \centering
         \includegraphics[width=\textwidth]{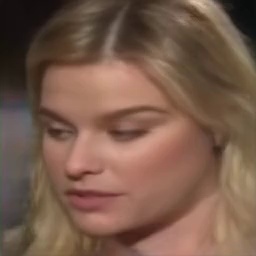}
         \par
     \end{minipage}
     \begin{minipage}[t]{0.19\textwidth}
         \centering
         \includegraphics[width=\textwidth]{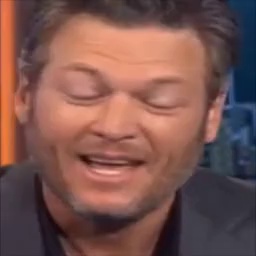}
         \par
     \end{minipage}
     \begin{minipage}[t]{0.19\textwidth}
         \centering
         \includegraphics[width=\textwidth]{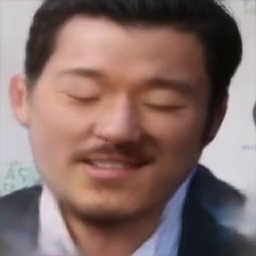}
         \par
     \end{minipage}
     \begin{minipage}[t]{0.19\textwidth}
         \centering
         \includegraphics[width=\textwidth]{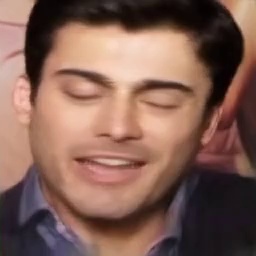}
         \par
     \end{minipage}
     \begin{minipage}[t]{0.19\textwidth}
         \centering
         \includegraphics[width=\textwidth]{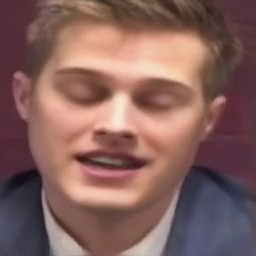}
         \par
     \end{minipage}
     \begin{minipage}[t]{0.19\textwidth}
         \centering
         \includegraphics[width=\textwidth]{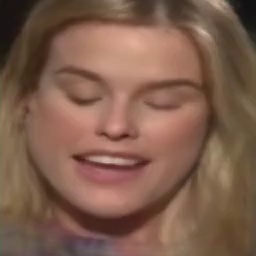}
         \par
     \end{minipage}
    \begin{minipage}[t]{0.19\textwidth}
         \centering
         \includegraphics[width=\textwidth]{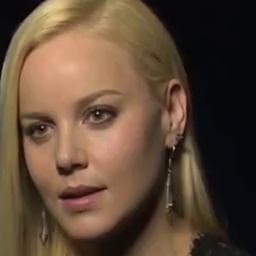}
         \par
     \end{minipage}
     \begin{minipage}[t]{0.19\textwidth}
         \centering
         \includegraphics[width=\textwidth]{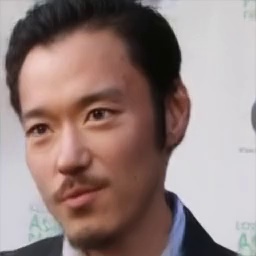}
         \par
     \end{minipage}
     \begin{minipage}[t]{0.19\textwidth}
         \centering
         \includegraphics[width=\textwidth]{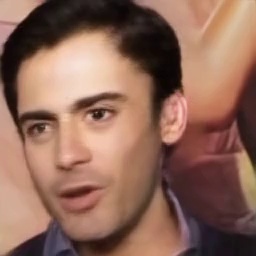}
         \par
     \end{minipage}
     \begin{minipage}[t]{0.19\textwidth}
         \centering
         \includegraphics[width=\textwidth]{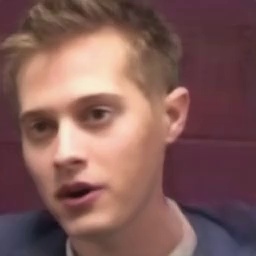}
         \par
     \end{minipage}
     \begin{minipage}[t]{0.19\textwidth}
         \centering
         \includegraphics[width=\textwidth]{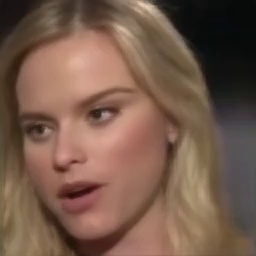}
         \par
     \end{minipage}
     \begin{minipage}[t]{0.19\textwidth}
         \centering
         \includegraphics[width=\textwidth]{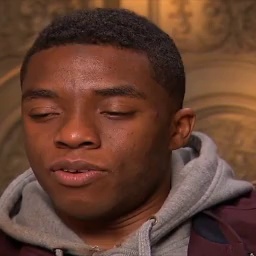}
         \par
     \end{minipage}
     \begin{minipage}[t]{0.19\textwidth}
         \centering
         \includegraphics[width=\textwidth]{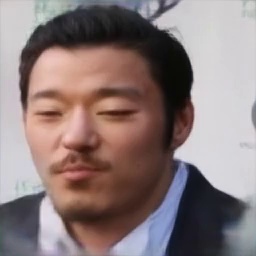}
         \par
     \end{minipage}
     \begin{minipage}[t]{0.19\textwidth}
         \centering
         \includegraphics[width=\textwidth]{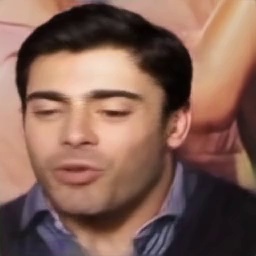}
         \par
     \end{minipage}
     \begin{minipage}[t]{0.19\textwidth}
         \centering
         \includegraphics[width=\textwidth]{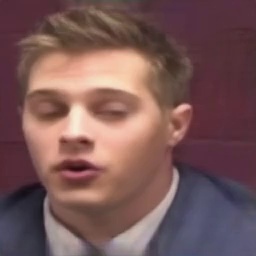}
         \par
     \end{minipage}
     \begin{minipage}[t]{0.19\textwidth}
         \centering
         \includegraphics[width=\textwidth]{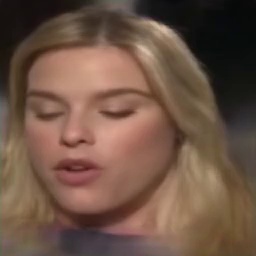}
         \par
     \end{minipage}
     \begin{minipage}[t]{0.19\textwidth}
         \centering
         \includegraphics[width=\textwidth]{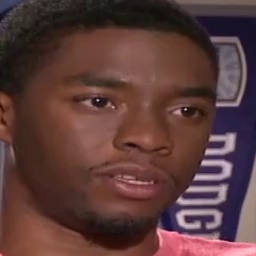}
         \par
     \end{minipage}
     \begin{minipage}[t]{0.19\textwidth}
         \centering
         \includegraphics[width=\textwidth]{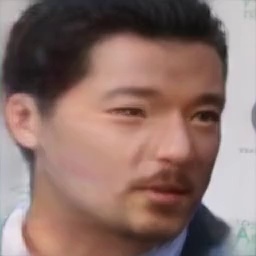}
         \par
     \end{minipage}
     \begin{minipage}[t]{0.19\textwidth}
         \centering
         \includegraphics[width=\textwidth]{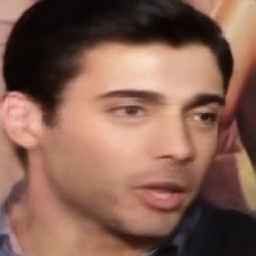}
         \par
     \end{minipage}
     \begin{minipage}[t]{0.19\textwidth}
         \centering
         \includegraphics[width=\textwidth]{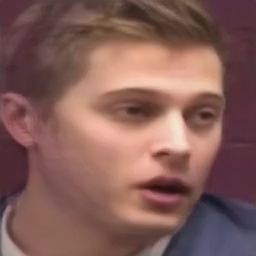}
         \par
     \end{minipage}
     \begin{minipage}[t]{0.19\textwidth}
         \centering
         \includegraphics[width=\textwidth]{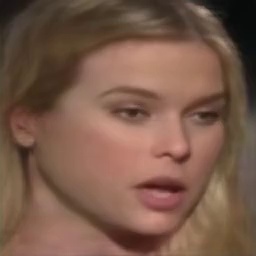}
         \par
     \end{minipage}
    \caption{\textbf{Training data examples.} The left-most column is the genuine video from $D_{gen}$ (last two examples are different videos of the same person), while the remaining four columns represent the corresponding forgeries from $D_{aug}$ generated with the top row as the target identity.}
    \label{appfig:training-dataset-examples}
\end{figure*}

\begin{figure*}[t]
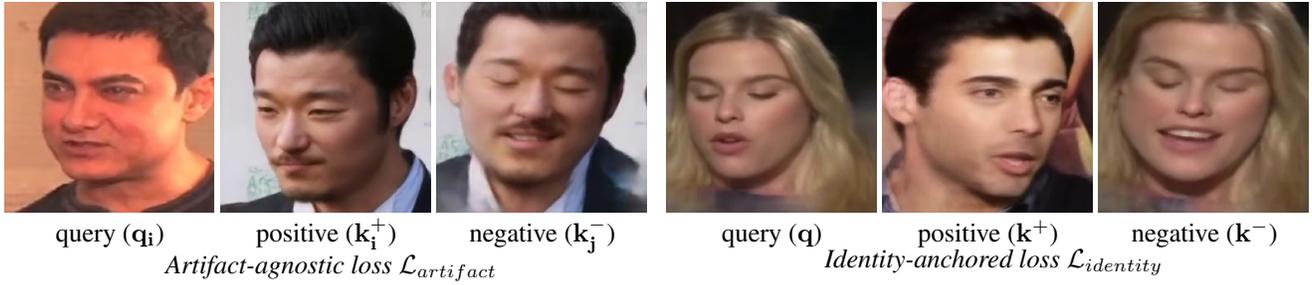

    \centering
    \begin{minipage}[t]{0.49\textwidth}
        \centering
         \begin{minipage}[t]{0.32\textwidth}
             \centering
             \includegraphics[width=\textwidth]{fig/training_00.jpg}
             query ($\mathbf{q_i}$)
             \par
         \end{minipage}
         \begin{minipage}[t]{0.32\textwidth}
             \centering
             \includegraphics[width=\textwidth]{fig/training_01.jpg}
             positive ($\mathbf{k_i^+}$)
             \par
         \end{minipage}
         \begin{minipage}[t]{0.32\textwidth}
             \centering
             \includegraphics[width=\textwidth]{fig/training_11.jpg}
             negative ($\mathbf{k_j^-}$)
             \par
         \end{minipage}
         \begin{minipage}[t]{0.01\textwidth}
             \phantom{\rule{0pt}{1em}}
         \end{minipage}
         \emph{Artifact-agnostic loss $\mathcal{L}_{artifact}$}
         \par
     \end{minipage}
     \begin{minipage}[t]{0.49\textwidth}
        \centering
         \begin{minipage}[t]{0.32\textwidth}
             \centering
             \includegraphics[width=\textwidth]{fig/training_24.jpg}
             query ($\mathbf{q}$)
             \par
         \end{minipage}
         \begin{minipage}[t]{0.32\textwidth}
             \centering
             \includegraphics[width=\textwidth]{fig/training_32.jpg}
             positive ($\mathbf{k^+}$)
             \par
         \end{minipage}
         \begin{minipage}[t]{0.32\textwidth}
             \centering
             \includegraphics[width=\textwidth]{fig/training_14.jpg}
             negative ($\mathbf{k^-}$)
             \par
        \end{minipage}
        \begin{minipage}[t]{0.01\textwidth}
        \phantom{\rule{0pt}{1em}}
        \end{minipage}
        \emph{Identity-anchored loss $\mathcal{L}_{identity}$}
        \par
     \end{minipage}
    \caption{\textbf{Contrastive pairs examples.} We present an illustrative selection of query, positive, and negative examples for the \emph{artifact-agnostic loss} ($\mathcal{L}_{artifact}$) and \emph{identity-anchored loss} ($\mathcal{L}_{identity}$). 
    }
    \label{appfig:contrastive-paris}
\end{figure*}

     Fig.~\ref{appfig:training-dataset-examples} presents sample frames from the training data. For each genuine video, we produce forgery results to create the augmented set $D_{aug}$. In particular, given the genuine example $V(s)\in D_{gen}$ and a target identity (portrait) $x$, we denote deepfake-augmented example $U\in D_{aug}$ as $U = \mathcal{F}_x\circ V(s)$. In Fig.~\ref{appfig:training-dataset-examples}, the left-most column illustrates video frames from each of the genuine examples $V(s)\in D_{gen}$ whereas the top row presents the target identities $x$. The remaining images in the grid depict $U = \mathcal{F}_x\circ V(s)$, whereby each image corresponds to the specific combination of $V(s)$ and $x$ from its respective row and column, respectively. Furthermore, Fig.~\ref{appfig:contrastive-paris} presents an illustrative selection of query, positive and negative examples of both the artifact-agnostic loss $\mathcal{L}_{artifact}$ and identity-anchored loss $\mathcal{L}_{identity}$, based on samples shown in Fig.~\ref{appfig:training-dataset-examples}.

\begin{figure*}[t]
    \centering
     \begin{minipage}[t]{0.24\textwidth}
         \centering
         \includegraphics[width=\textwidth]{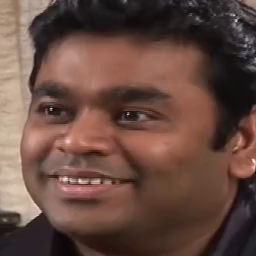}
         $D_{gen}$
         \par
     \end{minipage}
     \begin{minipage}[t]{0.24\textwidth}
         \centering
         \includegraphics[width=\textwidth]{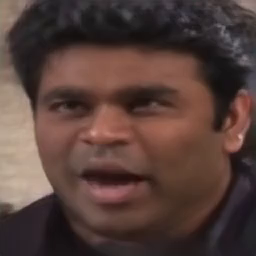}
         $D_{forg}$
         \par
     \end{minipage}
     \begin{minipage}[t]{0.24\textwidth}
         \centering
         \includegraphics[width=\textwidth]{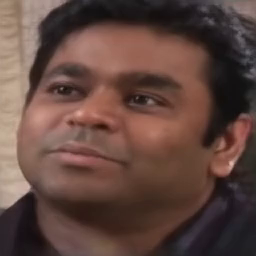}
         $D_{recon}$
         \par
     \end{minipage}\\
     \begin{minipage}[t]{0.24\textwidth}
         \centering
         \includegraphics[width=\textwidth]{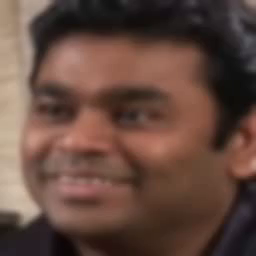}
         $GB(D_{gen})$
         \par
     \end{minipage}
     \begin{minipage}[t]{0.24\textwidth}
         \centering
         \includegraphics[width=\textwidth]{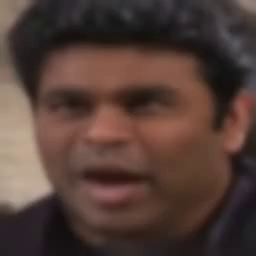}
         $GB(D_{forg})$
         \par
     \end{minipage}
     \begin{minipage}[t]{0.24\textwidth}
         \centering
         \includegraphics[width=\textwidth]{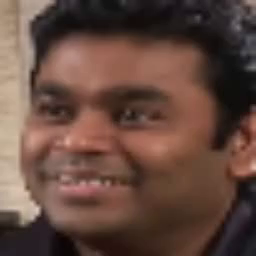}
         $JPEG(D_{gen})$
         \par
     \end{minipage}
     \begin{minipage}[t]{0.24\textwidth}
         \centering
         \includegraphics[width=\textwidth]{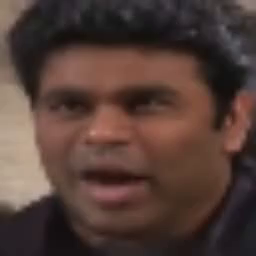}
         $JPEG(D_{forg})$
         \par
     \end{minipage}\\
     \begin{minipage}[t]{0.24\textwidth}
         \centering
         \includegraphics[width=\textwidth]{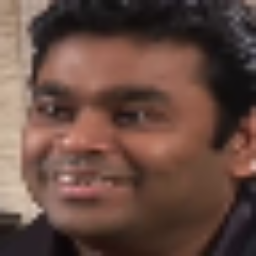}
         $Resize(D_{gen})$
         \par
     \end{minipage}
     \begin{minipage}[t]{0.24\textwidth}
         \centering
         \includegraphics[width=\textwidth]{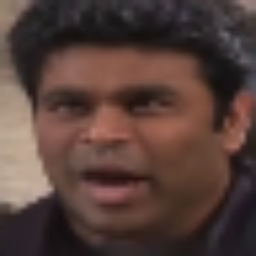}
         $Resize(D_{forg})$
         \par
     \end{minipage}
     \begin{minipage}[t]{0.24\textwidth}
         \centering
         \includegraphics[width=\textwidth]{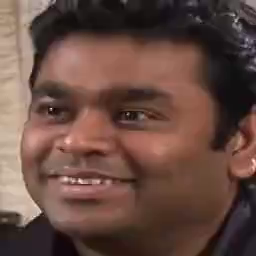}
         $VC(D_{gen})$
         \par
     \end{minipage}
     \begin{minipage}[t]{0.24\textwidth}
         \centering
         \includegraphics[width=\textwidth]{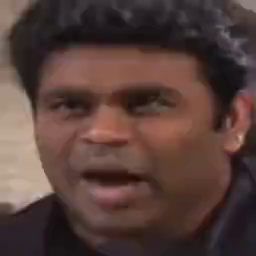}
          $VC(D_{forg})$
         \par
     \end{minipage}
    \caption{\textbf{Testing data examples of FR.} The data preview under different protocols.}
    \label{appfig:FR-testing-dataset-examples}
\end{figure*}

\begin{figure*}[t]
    \centering
     \begin{minipage}[t]{0.24\textwidth}
         \centering
         \includegraphics[width=\textwidth]{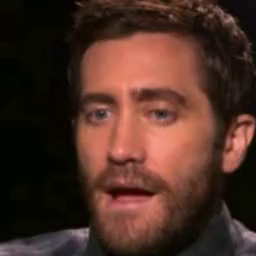}
         $D_{gen}$
         \par
     \end{minipage}
     \begin{minipage}[t]{0.24\textwidth}
         \centering
         \includegraphics[width=\textwidth]{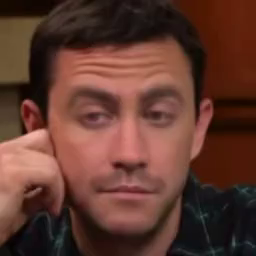}
         $D_{forg}$
         \par
     \end{minipage}
     \begin{minipage}[t]{0.24\textwidth}
         \centering
         \includegraphics[width=\textwidth]{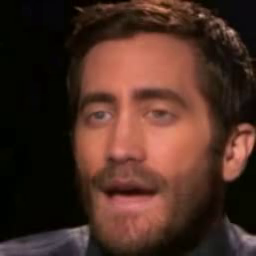}
         $D_{recon}$
         \par
     \end{minipage}\\
     \begin{minipage}[t]{0.24\textwidth}
         \centering
         \includegraphics[width=\textwidth]{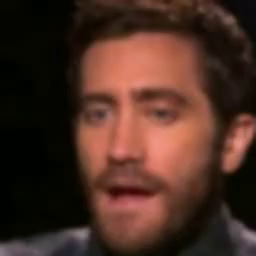}
         $GB(D_{gen})$
         \par
     \end{minipage}
     \begin{minipage}[t]{0.24\textwidth}
         \centering
         \includegraphics[width=\textwidth]{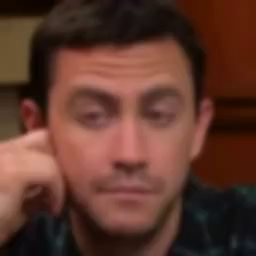}
         $GB(D_{forg})$
         \par
     \end{minipage}
     \begin{minipage}[t]{0.24\textwidth}
         \centering
         \includegraphics[width=\textwidth]{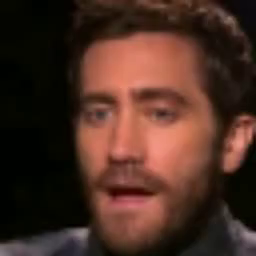}
         $JPEG(D_{gen})$
         \par
     \end{minipage}
     \begin{minipage}[t]{0.24\textwidth}
         \centering
         \includegraphics[width=\textwidth]{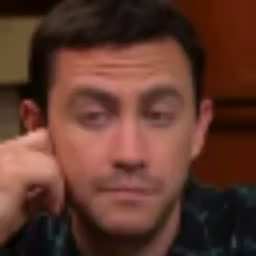}
         $JPEG(D_{forg})$
         \par
     \end{minipage}\\
     \begin{minipage}[t]{0.24\textwidth}
         \centering
         \includegraphics[width=\textwidth]{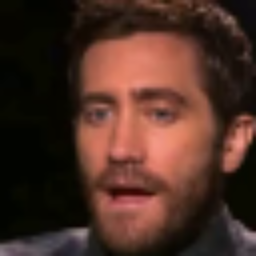}
         $Resize(D_{gen})$
         \par
     \end{minipage}
     \begin{minipage}[t]{0.24\textwidth}
         \centering
         \includegraphics[width=\textwidth]{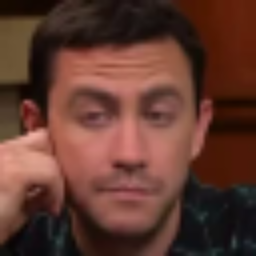}
         $Resize(D_{forg})$
         \par
     \end{minipage}
     \begin{minipage}[t]{0.24\textwidth}
         \centering
         \includegraphics[width=\textwidth]{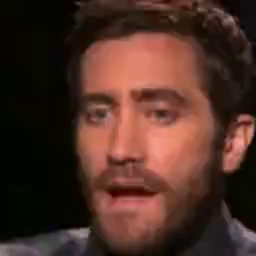}
         $VC(D_{gen})$
         \par
     \end{minipage}
     \begin{minipage}[t]{0.24\textwidth}
         \centering
         \includegraphics[width=\textwidth]{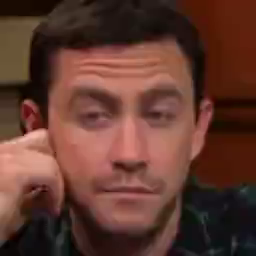}
          $VC(D_{forg})$
         \par
     \end{minipage}
    \caption{\textbf{Testing data examples of FS.} The data preview under different protocols.}
    \label{appfig:FS-testing-dataset-examples}
\end{figure*}
    \paragraph{Testing environment: conventional and \protocolAbbrv.}
    
    Test examples under the evaluation protocols for the two subdivisions, FR and FS, are illustrated in Fig.~\ref{appfig:FR-testing-dataset-examples} and Fig.~\ref{appfig:FS-testing-dataset-examples} respectively. In these figures, we display instances corresponding to $D_{gen}$, $D_{forg}$, $D_{recon}$, along with $\tilde{D}_{gen}$ and $\tilde{D}_{forg}$. Here, $\tilde{D}_{gen/forg}$ is depicted as $\text{f.}(D_{gen/forg})$, where the surrogate function \text{f.} is Resize, JPEG compression, Video compression (VC), or Gaussian blur (GB).

    Under the \pcolconventional protocol, the differentiation is made between $D_{gen}$ and $D_{forg}$, which often demonstrates a perceptible distribution shift in terms of image sharpness. Contrastingly, the \pcolwhite protocol differentiates between $D_{recon}$ and $D_{forg}$, while \pcolsur makes the distinction between $\tilde{D}_{gen}$ and $\tilde{D}_{forg}$, both of which typically exhibit similar sharpness levels. This illustrates the nuances and intricacies inherent in deepfake detection across different evaluation protocols.


\end{document}